\documentclass{article}

\usepackage[main, final]{neurips_2026}

\usepackage{graphicx}
\usepackage[utf8]{inputenc} % allow utf-8 input
\usepackage[T1]{fontenc}    % use 8-bit T1 fonts
\usepackage{hyperref}       % hyperlinks
\usepackage{url}            % simple URL typesetting
\usepackage{booktabs}       % professional-quality tables
\usepackage{amsfonts}       % blackboard math symbols
\usepackage{nicefrac}       % compact symbols for 1/2, etc.
\usepackage{microtype}      % microtypography
\usepackage{xcolor}         % colors
\usepackage[table]{xcolor}
\usepackage{caption}
\usepackage{makecell}
\usepackage{amsmath}
\usepackage[most]{tcolorbox}
\usepackage{fancyvrb}
\newcommand{\gain}[1]{\textcolor{green!60!black}{\scalebox{0.75}{\tiny{$+$#1}}}}

\newcommand{\method}{\textsc{PERIA}}
\newcommand{\methodfull}{\textit{PERception-Interaction-reason Agent} (\textbf{\method})}

\title{Perceive, Interact, Reason: Building Tool-Augmented Visual Agents for Spatial Reasoning}

\author{%
Changye Li$^{1}$,
Meng Lu$^{2}$,
Yi Wu$^{1}$,
Ligeng Zhu$^{3}$\\
$^{1}$Institute for Interdisciplinary Information Sciences, Tsinghua University\\
$^{2}$Virginia Tech \quad
$^{3}$NVIDIA\\\\
\textbf{Code:} \url{https://github.com/antoinegg1/PEDIA}\\
\textbf{Model:} \url{https://huggingface.co/Changyeli03/pedia_model}\\
\textbf{Dataset:} \url{https://huggingface.co/datasets/Changyeli03/pedia_data}
}

\begin{document}

\addtocontents{toc}{\protect\setcounter{tocdepth}{-1}}

\maketitle

\begin{abstract}
While recent vision-language models (VLMs) demonstrate strong multimodal understanding, they remain limited in spatial reasoning tasks that require active evidence acquisition and multi-step visual interaction. This limitation suggests that relying solely on implicit visual representations from vision encoders is insufficient for recovering fine-grained spatial evidence. We introduce \methodfull, a tool-augmented visual agent for spatial reasoning tasks across map reasoning, visual probing, and vision reconstruction. \method\ uses two lightweight tool families: \textit{vision perception tools} for exposing textual, symbolic, and spatial evidence, and \textit{vision interaction tools} for manipulating visual context, tracing paths, and verifying spatial relations. To train \method, we develop a unified recipe that combines supervised tool-use trajectory synthesis, composite rewards, and Observation-Relaxed Group-in-Group Policy Optimization (OR-GIGPO) for effective multi-tool behavior. Experiments on 13 benchmarks from 8 datasets show that \method-8B improves over the Qwen3-8B backbone by $10.0$ percentage points on in-distribution benchmarks and $4.4$ percentage points on out-of-distribution benchmarks, while outperforming previous state-of-the-art baselines of similar size by $6.9$--$14.8$ percentage points. It also achieves performance comparable to much larger models such as Qwen3-VL-235B-A22B-Thinking and GPT-5, demonstrating the effectiveness of \method\ in enhancing spatial reasoning capabilities.
\end{abstract}

\section{Introduction}

Vision-language models (VLMs) have made strong progress in multimodal understanding and vision-language reasoning~\citep{bai2025qwen3vltechnicalreport,comanici2025gemini,wang2025internvl35advancingopensourcemultimodal,singh2025openai}. 
Chain-of-thought prompting~\citep{wei2023chainofthoughtpromptingelicitsreasoning} and reasoning-oriented post training~\citep{zhang2024multimodalchainofthoughtreasoninglanguage} further enhance their ability to perform visual reasoning. 
However, many complex visual tasks require more than reasoning over a static visual representation, as models need to ground fine-grained evidence, extract task-relevant details, and reason about spatial relations through multiple intermediate steps~\citep{yue2024mmmumassivemultidisciplinemultimodal,ma20243dsrbench,stogiannidis2025mindgapbenchmarkingspatial}. 
Tool use provides a natural way to extend model capabilities. Inspired by LLM agents~\citep{yao2023reactsynergizingreasoningacting,schick2023toolformerlanguagemodelsteach}, recent VLMs have begun to use external visual tools for reasoning~\citep{openai2025o3o4mini,lu2025scalingagenticreinforcementlearning,fu2025refocusvisualeditingchain,wu2026vtoolr1vlmslearnthink}.

Existing tool-augmented VLMs show promising gains on structured visual tasks such as chart understanding and mathematical problem solving, where answers often come from explicit evidence and mature expert tools such as ChartMoE~\citep{xu2025chartmoemixturediverselyaligned}. 
Yet many spatial tasks cannot be solved by reading a local text span, chart element, equation, or detected object. 
For example, subway-map reasoning requires aligning station labels with route topology~\citep{feng2025can,feng2025rewardmap}, while MapTrace requires drawing valid paths over maps~\citep{panagopoulou2025maptracescalabledatageneration}. 
These tasks require a coherent spatial state that links entities across regions and supports topology-dependent reasoning.

The main limitation is a mismatch between current visual tools and spatial reasoning. 
Most tools read evidence or focus regions through OCR or cropping, but they do not by themselves build or update a spatial state across tool calls. 
Existing visual-tool methods also use limited tool spaces or domain-specific training data, which limits transfer to broader spatial reasoning~\citep{fu2025refocusvisualeditingchain,lai2025minio3scalingreasoningpatterns}. 
Raw tool access is not enough either: Section~\ref{sec:motivation} shows that tools provide useful evidence, but open-source VLMs often fail to use them reliably without tool-use training. 
Thus, the root issue is the lack of a trained policy that can compose perception and interaction tools, gather local evidence, recover global structure, and reason over accumulated observations.

\begin{figure}[t]
    \centering
    \includegraphics[width=\linewidth]{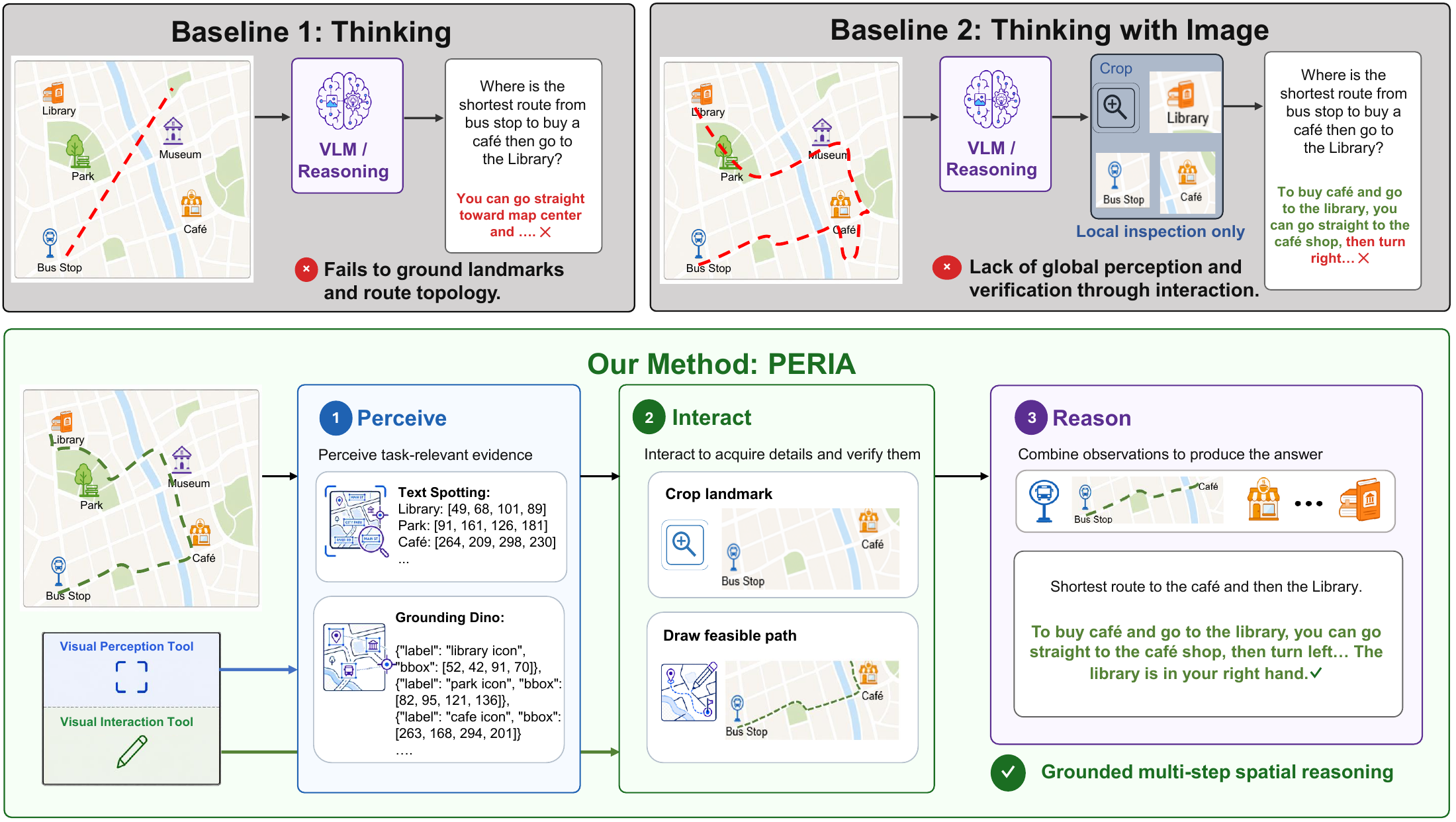}
    \caption{
    \textbf{Overview of our proposed \methodfull.}
    \methodfull\ first uses perception tools to gather global spatial evidence, then applies interaction tools for fine-grained local analysis, and finally reasons over accumulated observations. This design addresses the limits of conventional methods, which either miss local visual details or fail to recover global spatial structure.
    }
    \label{fig:framework_overview}
\end{figure}

This motivates our central question: \emph{How can we train visual agents to leverage diverse visual tools within a unified reasoning process for spatial reasoning tasks?}
We introduce \methodfull, a tool-augmented visual agent for spatial reasoning tasks. As shown in Figure~\ref{fig:framework_overview}, \method\ organizes visual problem solving into a unified process: the agent perceives task-relevant evidence, interacts with visual content for spatial planning and verification, and reasons over accumulated observations to produce the answer. It is supported by two complementary tool families: \emph{vision perception tools} and \emph{vision interaction tools}. For policy optimization, we further develop \emph{Observation-Relaxed Group-in-Group Policy Optimization} (OR-GIGPO) to assign credit across multi-step visual tool-use trajectories. Empirically, \method-8B surpasses previous state-of-the-art baselines of similar size and reaches performance comparable to current leading models. Our contributions are summarized as follows:
\begin{itemize}
    \item We introduce a diverse perception--interaction toolset for spatial reasoning and formulate \method\ as a general framework for tool-augmented visual agents that reason, plan, and verify over images. We further synthesize a diverse, high-quality trajectory dataset for supervised fine-tuning, enabling reliable tool invocation and tool-grounded reasoning.

    \item We develop a unified reinforcement training recipe centered on \emph{Observation-Relaxed Group-in-Group Policy Optimization} (OR-GIGPO), our core optimization method for visual tool-use trajectories. Together with task-specific rewards, OR-GIGPO enables fine-grained credit assignment and effective multi-tool composition. Ablation studies show that OR-GIGPO surpasses mainstream reinforcement learning algorithms in our tasks.

    \item We instantiate \method\ as a spatial reasoning agent across diverse task families, including map reasoning, visual probing, and vision reconstruction. Across 13 benchmarks from 8 datasets, \method-8B achieves consistent gains of $6.9$--$14.8$ percentage points over strong VLM and tool-augmented baselines at a comparable scale, while approaching the performance of much larger models such as Qwen3-VL-235B-A22B-Thinking and GPT-5.
\end{itemize}

\section{Related Work}
\label{gen_inst}

\paragraph{Tool-Augmented Visual Reasoning}
Tool use is a common way to extend language agents beyond text generation. ReAct~\citep{yao2023reactsynergizingreasoningacting} shows that models can interleave reasoning traces with external actions, forming a simple recipe for multi-step problem solving and environment feedback integration. Recent VLM agents extend visual reasoning with tool calls in domains such as charts, mathematics, and tables, helping models read text, edit images, or call expert modules with strong empirical gains~\citep{lu2025scalingagenticreinforcementlearning,fu2025refocusvisualeditingchain,wu2026vtoolr1vlmslearnthink,zhao2025pyvisionagenticvisiondynamic}. 
Another line of work explores interactive visual tools, where the DeepEyes series and Mini-o3 demonstrate the effectiveness of multi-turn image cropping, and Think3D studies 3D manipulation tools for spatial exploration~\citep{zheng2026deepeyesincentivizingthinkingimages,hong2026deepeyesv2agenticmultimodalmodel,lai2025minio3scalingreasoningpatterns,zhang2026think3dthinkingspacespatial}. 
However, the former line leaves spatial reasoning underexplored, while the latter often focuses on a small toolset or a narrow task family. In contrast, our work jointly trains perception and interaction tools within a unified framework, targeting general spatial reasoning agents that can read, inspect, trace, verify, and compose evidence across diverse task families.

\paragraph{Spatial Reasoning in VLMs.}
Spatial reasoning tests whether VLMs can connect visual evidence with relations, routes, layouts, and geometric changes. Prior work studies this ability across domains such as 3D spatial understanding~\citep{chen2024spatialvlmendowingvisionlanguagemodels,ma20243dsrbench}, map and geospatial reasoning~\citep{dihan2025mapevalmapbasedevaluationgeospatial,xing2025largevisionlanguagemodels,feng2026reasonmapfinegrainedvisualreasoning}, and visual search or probing in cluttered images~\citep{wu2023vguidedvisualsearch,lai2025minio3scalingreasoningpatterns,gavrikov2025visualoverloadprobingvisualunderstanding}. Recent work further moves from isolated tasks toward broader spatial intelligence, including shared spatial skills, low-level visual primitives, simulation, and reconstruction~\citep{stogiannidis2025mindgapbenchmarkingspatial,jia2026omnispatialcomprehensivespatialreasoning,chen2026babyvisionvisualreasoninglanguage,wu2026visual}. Together, these studies suggest that spatial reasoning requires global-local alignment, path tracing, relation checking, and multi-step planning, motivating our perception--interaction tool design that combines visual evidence acquisition with explicit visual interaction.

\paragraph{Reinforcement Learning for Tool-Use Agents.} Reinforcement learning (RL) is widely used in post-training: PPO~\citep{schulman2017proximalpolicyoptimizationalgorithms} provides a general policy optimization framework, GRPO~\citep{shao2024deepseekmathpushinglimitsmathematical} estimates relative advantages from groups of sampled responses, and DAPO~\citep{yu2025dapoopensourcellmreinforcement} improves large-scale RL with decoupled clipping and dynamic sampling. For tool-use agents, sparse rewards and long-horizon interactions make it difficult to evaluate intermediate decisions; RAGEN addresses this issue in multi-turn agent RL, while GIGPO assigns learning signals to both full trajectories and intermediate states~\citep{wang2025ragenunderstandingselfevolutionllm,feng2025groupingrouppolicyoptimizationllm}. Our setting further introduces visual observation variability from tool-agent outputs, motivating observation-relaxed policy optimization instead of exact state matching.

\section{Motivation: The Gap Between Having Tools and Using Tools}
\label{sec:motivation}
\begin{figure}[h]
  \centering

  \begin{minipage}[t]{0.47\linewidth}
  \vspace{0pt}
    \centering
    \includegraphics[width=\linewidth]{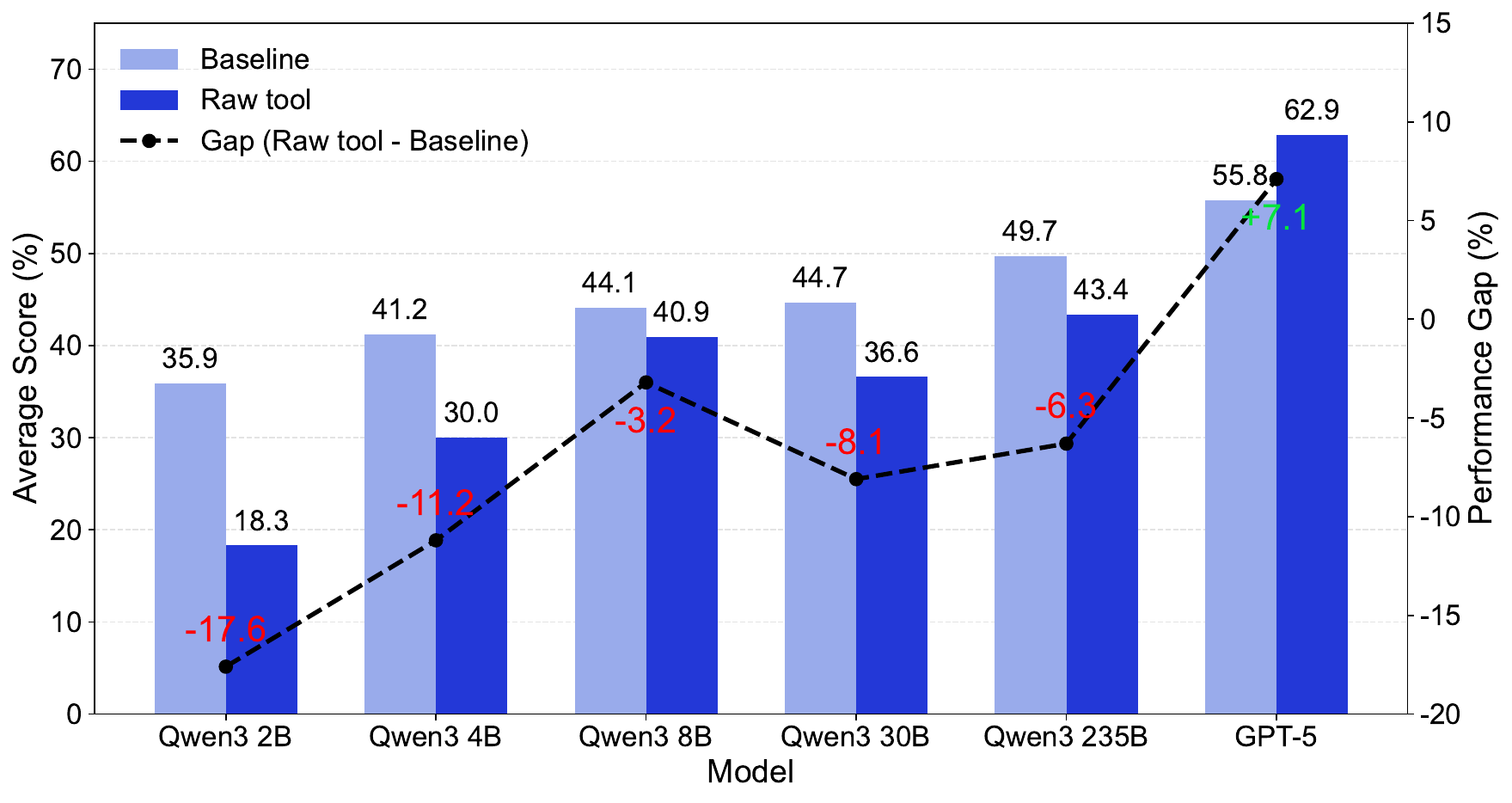}
    \captionof{figure}{\textbf{Tool access does not guarantee tool competence.}
We compare reasoning without tools (\textit{Baseline}) with direct tool-augmented reasoning (\textit{Raw tool}).
GPT-5 benefits from raw tool access, but Qwen3-VL-Thinking models, abbreviated as Qwen3 in the figure, often degrade without tool-use training.
The gap narrows with model size but remains clear.}
    \label{fig:model-tool-gap}
  \end{minipage}
  \hfill
  \begin{minipage}[t]{0.50\linewidth}
  \vspace{0pt}
    \centering
    \footnotesize
    \resizebox{\linewidth}{!}{%
    \begin{tabular}{lccccc}
      \toprule
     \textbf{Model} & \textbf{\makecell{Error\\Count}} & \textbf{\makecell{Tool-Call\\Omission}} & \textbf{\makecell{Tool-Induced\\Error}} & \textbf{\makecell{Format Error}} & \textbf{Other} \\
      \midrule
      \texttt{Qwen3-2B}   & 296 (98.7\%) & \textbf{55.9\%} & \underline{31.1\%} & 2.0\% & 11.0\% \\
      \texttt{Qwen3-4B}   & 176 (58.7\%) & \textbf{71.6\%} & \underline{25.0\%} & 2.3\% & 1.1\% \\
      \texttt{Qwen3-8B}   & 166 (55.4\%) & \textbf{59.0\%} & \underline{30.7\%} & 1.8\% & 8.4\% \\
      \texttt{Qwen3-30B-A3B}  & 172 (57.3\%) & \textbf{68.6\%} & \underline{30.8\%} & 0.6\% & 0.0\% \\
      \texttt{Qwen3-235B} & 133 (44.3\%) & \underline{38.3\%} & \textbf{60.9\%} & 0.8\% & 0.0\% \\
      \bottomrule
    \end{tabular}
    }
    \captionof{table}{\textbf{Most failures come from not using tools well.}
We analyze 300 cases from 6 benchmarks where GPT-5 solves the task with tools but Qwen3-VL-Thinking models, abbreviated as Qwen3 in the table, fail in most cases.
Tool-call omission and tool-induced errors dominate, showing that current models mainly struggle with tool selection, invocation, and feedback integration.
Larger models fail less often, but clear tool-use gaps remain.}
    \label{tab:error-patterns}
  \end{minipage}
\end{figure}

\textbf{\textit{Having visual tools does not mean using them well.}} We test this gap on 6 benchmarks from 4 datasets: MapTrace~\citep{panagopoulou2025maptracescalabledatageneration}, ReasonMap~\citep{feng2025can}, ReasonMap-Plus~\citep{feng2025rewardmap}, and Visual Probing Easy/Medium/Hard~\citep{lai2025minio3scalingreasoningpatterns}. The raw-tool setting gives each model access to 18 perception and interaction tools for evidence extraction, spatial annotation, and local inspection; Section~\ref{sec:overall_architecture} details the sandbox. As shown in Figure~\ref{fig:model-tool-gap}, GPT-5 improves with these tools, confirming that they expose useful spatial evidence. In contrast, Qwen3-VL-Thinking models often degrade, showing that open-source VLMs do not reliably turn tool access into better spatial reasoning without dedicated tool-use training.

We further diagnose the gap by inspecting 300 cases where GPT-5 succeeds with tools but Qwen3-VL-Thinking models fail. Table~\ref{tab:error-patterns} shows that most errors are semantic tool-use failures rather than formatting failures. Models often skip needed tools or are misled by tool outputs during later reasoning. Thus, the gap is mainly a policy-learning problem: models must learn when to call tools, which tools to call, and how to integrate tool feedback. This motivates \method, which trains agents to coordinate perception and interaction tools throughout multi-step spatial reasoning.

\section{Methodology}
\label{methodology}

\begin{figure}[h]
    \centering
    \includegraphics[width=\linewidth]{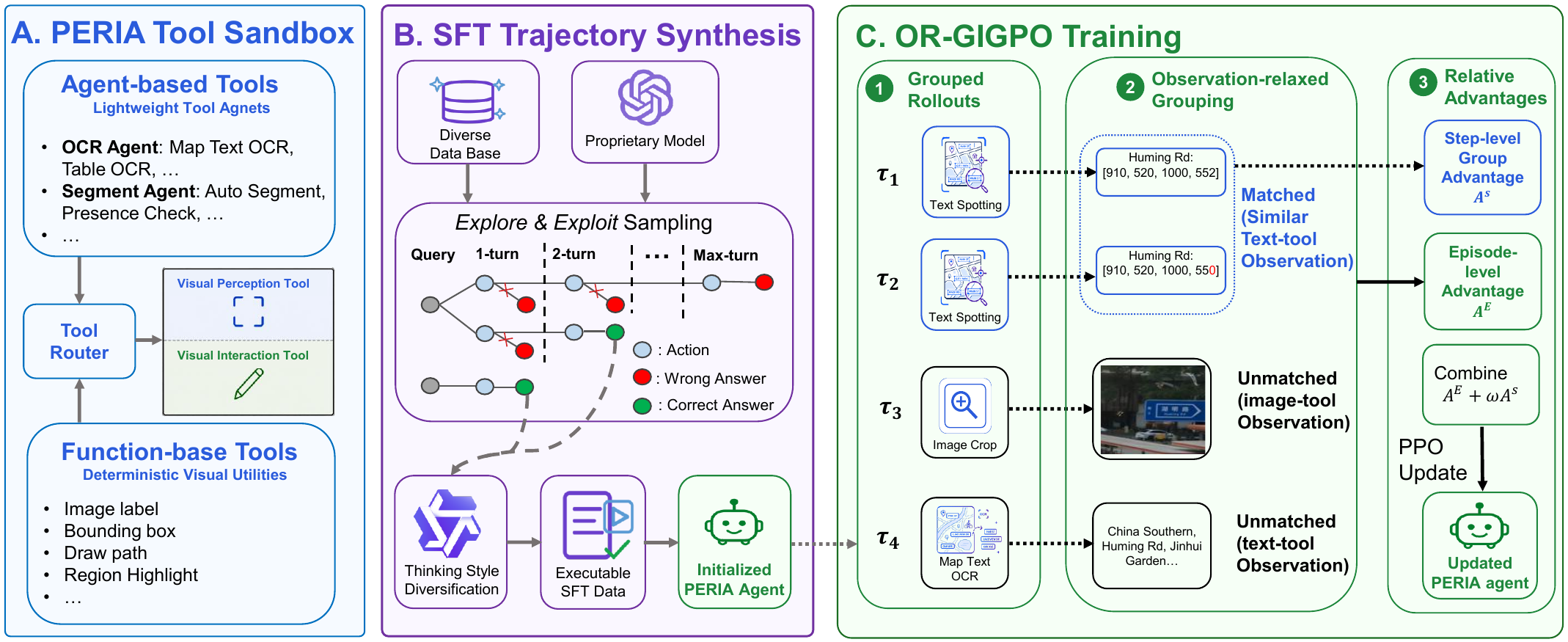}
    \caption{
\textbf{\method\ learns tool use through sandbox construction, trajectory synthesis, and OR-GIGPO.}
The pipeline first builds the \method\ Tool Sandbox, which organizes function-based visual utilities and lightweight agent-based tools into perception and interaction tools.
It then synthesizes executable SFT trajectories with budgeted explore-and-exploit sampling and thinking-style diversification.
Finally, OR-GIGPO groups rollouts by relaxed observations and combines step-level and episode-level advantages to assign credit across multi-step tool-use trajectories.
}
    \label{fig:method_overview}
\end{figure}

In this section, we describe the construction of \method, as illustrated in Figure~\ref{fig:method_overview}. The overall design follows three stages. First, we formulate \method\ as a tool-augmented visual agent and build a modular tool sandbox that exposes both deterministic visual utilities and lightweight tool agents, organized into perception and interaction tools for acquiring evidence and manipulating visual context. Second, we synthesize executable tool-use trajectories for supervised fine-tuning (SFT): a proprietary model explores candidate tool-call paths under increasing turn budgets, and an expert model diversifies the intermediate reasoning styles. Third, we train the initialized agent with \emph{Observation-Relaxed Group-in-Group Policy Optimization} (OR-GIGPO), which groups rollouts by relaxed observations and combines episode-level and step-level advantages for multi-step tool-use credit assignment. We detail the \method\ formulation and tool sandbox in Section~\ref{sec:overall_architecture}, trajectory synthesis in Section~\ref{sec:pir_trajectory_synthesis}, and OR-GIGPO training with composite rewards in Section~\ref{sec:training_recipe}.

\subsection{\method\ Formulation and Tool Sandbox}
\label{sec:overall_architecture}

\paragraph{\method\ as a POMDP.}
Following prior formulations of tool-integrated and agentic reasoning~\citep{yao2023reactsynergizingreasoningacting, lu2025scalingagenticreinforcementlearning}, we formulate \method\ as a partially observable Markov decision process (POMDP) $\langle \mathcal{S}, \mathcal{O}, \mathcal{A}, \mathcal{I}, \mathcal{P}, \mathcal{R} \rangle$, where $\mathcal{I}$, $\mathcal{S}$, $\mathcal{O}$, $\mathcal{A}$, $\mathcal{P}$, and $\mathcal{R}$ denote the instruction, latent state, observation, action, transition, and reward spaces. Given an instruction $x \in \mathcal{I}$ and an initial visual observation $o_0$, we denote the task input as $q=(x,o_0)$. The agent iteratively generates a reasoning thought and an action:
\[
g_t \sim \pi_\theta(\cdot \mid q,c_{t-1}), \qquad
a_t \sim \pi_\theta(\cdot \mid q,c_{t-1},g_t),
\]
where $c_{t-1}=(g_1,a_1,o_1,\ldots,g_{t-1},a_{t-1},o_{t-1})$ is the interaction history, and $a_t$ can be a perception-tool call, an interaction-tool call, or a final-answer action. Tool actions update the environment through $\mathcal{P}$ and return new observations, which are appended to the history. The resulting trajectory is
\[
\tau=(g_1,a_1,o_1,\ldots,g_{T-1},a_{T-1},o_{T-1},g_T,y)\sim\pi_\theta(\tau\mid q),
\]
where $y=a_T$ is the final-answer action generated after the final reasoning step $g_T$.

\paragraph{\method\ Tool Sandbox.}
To support flexible tool-augmented spatial reasoning, we build a modular \textit{\method\ Tool Sandbox} with 18 tools, implemented through function-based utilities and 3 lightweight tool agents. The sandbox provides a unified and extensible interface for tool invocation. We organize the tools into two categories: \textit{visual perception tools}, which extract task-relevant evidence such as text, coordinates, and object presence, represented by \texttt{text\_spotting}, \texttt{map\_text\_ocr}, and \texttt{auto\_segment}; and \textit{visual interaction tools}, which explicitly manipulate or annotate the visual scene, represented by \texttt{draw\_path}, \texttt{bounding\_box}, and \texttt{image\_crop}. Our sandbox design draws inspiration from recent tool-augmented systems~\citep{wu2026vtoolr1vlmslearnthink, fu2025refocusvisualeditingchain, lu2025scalingagenticreinforcementlearning}, while introducing newly designed and task-specific tools tailored to spatial reasoning. Full tool definitions and sandbox design are provided in Appendix~\ref{app:tool_sandbox}.

\subsection{Synthesizing Tool-Use Trajectories}
\label{sec:pir_trajectory_synthesis}

Our preliminary analysis in Section~\ref{sec:motivation} shows that directly enabling tool calls does not consistently improve model performance. We therefore synthesize tool-executing trajectories with a proprietary model (\texttt{GPT-5}) as supervised fine-tuning data to enhance the tool-calling capability of models.

Given a training instance $(q,y^\ast)$, where $q=(x,o_0)$, we generate trajectories with an \textit{explore-and-exploit} strategy. At round $t$, $\pi_{\rm pro}$ expands each current prefix by one reasoning--action step:
\[
\{\tau_t^{(i)}=(\tau_{t-1}^{(i)}, g_t^{(i)}, a_t^{(i)})\}_{i=1}^{n_{\rm sample}}
\sim \pi_{\rm pro}(\cdot \mid q,c_{t-1}^{(i)}),
\]
where $a_t^{(i)}$ is either a tool call or a final-answer action; if it is a tool call, the returned observation $o_t^{(i)}$ is appended to the next prefix, while final-answer actions are denoted by $y_t^{(i)}$.
We accept a trajectory once $y_t^{(i)}=y^\ast$; otherwise, we add error feedback and continue expansion until a valid trajectory is found or the turn budget $T$ is reached.

To diversify reasoning, we rewrite only the intermediate thoughts $\{g_t\}$ with an open-weight expert model (\texttt{Qwen3-VL-235B-A22B-Thinking}), while keeping tool calls, observations, and answers fixed. Full details of SFT trajectory synthesis are provided in Appendix~\ref{app:Trajectory_Synthesis}.

\subsection{Training Tool-Augmented Spatial Agents}
\label{sec:training_recipe}

We first fine-tune the model on the trajectories synthesized in Section~\ref{sec:pir_trajectory_synthesis}, which equips the agent with basic tool-use abilities and the Perceive--Interact--Reason workflow. Since SFT mainly imitates existing trajectories and is insufficient for flexible tool composition, we further optimize the agent with reinforcement learning. To better handle multi-tool trajectories, we design OR-GIGPO for credit assignment and use a composite reward for spatial reasoning.

\paragraph{OR-GIGPO.}
Motivated by Group-in-Group Policy Optimization (GIGPO)~\citep{feng2025groupingrouppolicyoptimizationllm}, we design \textit{Observation-Relaxed Group-in-Group Policy Optimization} (OR-GIGPO) for multi-step spatial reasoning training. GIGPO combines episode-level and step-level advantages, comparing actions taken from the same anchor state. OR-GIGPO instead groups actions by their resulting tool observations within the same task. Since equivalent tool calls may return slightly different outputs, especially from tool agents, we relax exact observation matching to accommodate these variations.

OR-GIGPO retains the two-level advantage structure but uses action outcomes for step-level grouping. Following Section~\ref{sec:overall_architecture}, let $q=(x,o_0)$ denote the instruction and initial visual observation. We define the episode-level rollout group and the step-level outcome group as
\[
\mathcal{G}_E(q)=\{\tau_i:q_i=q\}_{i=1}^{N}, 
\mathcal{G}_S(q,\tilde o)=
\bigl\{
(a_t^{(i)},R_t^{(i)}):
q_i=q,\
\underbrace{
h(o_t^{(i)})=h(\tilde o)
\lor
{\rm sim}(o_t^{(i)},\tilde o)\geq\delta
}_{\text{observation-relaxed matching}}
\bigr\}.
\]
Here $\tau_i$ is the $i$-th rollout, $a_t^{(i)}$ is the action generated conditioned on $(q^{(i)},c_{t-1}^{(i)},g_t^{(i)})$, and $o_t^{(i)}$ is the observation returned after executing that action. 
For step-level credit assignment, we associate each action with its discounted return $R_t^{(i)}=\sum_{k=t}^{T}\gamma^{k-t}r_k^{(i)}$, where $T$ is the number of turns, $\gamma$ is the discount factor, and $r_k^{(i)}$ is the reward at turn $k$. 
The episode group $\mathcal{G}_E(q)$ contains rollouts from the same input, while $\mathcal{G}_S(q,\tilde o)$ groups action-return pairs by matching their resulting observations to $\tilde o$, without requiring identical pre-action states. 
Here $h(\cdot)$ is used for exact observation matching, and ${\rm sim}(\cdot,\cdot)$ matches semantically similar textual observations from tool agents, implemented with Python's \texttt{SequenceMatcher}.

For each rollout $\tau_i$, let $R_i=R(\tau_i)$ be the trajectory-level reward. We compute the episode-level advantage $A_E$ and the step-level advantage $A_S$ as
\[
A_E(\tau_i)=\frac{R_i-\mu_E(q)}{\sigma_E(q)+\epsilon_{\rm adv}}, \qquad
A_S(a_t^{(i)})=\frac{R_t^{(i)}-\mu_S(q,\tilde o)}{\sigma_S(q,\tilde o)+\epsilon_{\rm adv}}.
\]
Here $\mu_E(q)$ and $\sigma_E(q)$ are the mean and standard deviation of trajectory rewards over rollouts from the same task input $q=(x,o_0)$, and $\epsilon_{\rm adv}$ is a small constant for numerical stability. 
The step-level statistics $\mu_S(q,\tilde o)$ and $\sigma_S(q,\tilde o)$ are computed over discounted returns in the same outcome group $\mathcal{G}_S(q,\tilde o)$. 
In our trajectory-level reward setting, this step-level return propagates the final rollout reward back to intermediate tool decisions.

The final advantage combines trajectory-level credit and outcome-group credit:
\[
A_t^{(i)}=A_E(\tau_i)+\omega A_S(a_t^{(i)}),
\]
where $\omega$ controls the strength of step-level credit assignment. We optimize the policy with a clipped objective using the combined advantage, without an explicit KL penalty:
\[
\mathcal{J}_{\rm OR\mbox{-}GIGPO}(\theta)
=
\mathbb{E}\left[
\frac{1}{\sum_{i=1}^{N}T_i}
\sum_{i=1}^{N}\sum_{\ell=1}^{T_i}
\min\left(
\rho_{\ell}^{(i)}\hat A_{\ell}^{(i)},
{\rm clip}(\rho_{\ell}^{(i)},1-\epsilon_{\rm clip},1+\epsilon_{\rm clip})\hat A_{\ell}^{(i)}
\right)
\right],
\]
where $\theta$ denotes the policy parameters and $\epsilon_{\rm clip}$ is the clipping threshold. Here $N$ is the number of rollouts, $T_i$ counts the generated thought and action tokens in rollout $i$, excluding tool observations, and $\hat A_{\ell}^{(i)}=A_t^{(i)}$ for a token generated at turn $t$. The token-level importance sampling ratio is
\[
\rho_{\ell}^{(i)}=
\frac{\pi_\theta(z_{\ell}^{(i)}\mid h_{\ell}^{(i)})}
{\pi_{\theta_{\rm old}}(z_{\ell}^{(i)}\mid h_{\ell}^{(i)})},
\]
where $z_{\ell}^{(i)}$ is the $\ell$-th generated token and $h_{\ell}^{(i)}$ is its full preceding context, including the input, previous generated tokens, and tool observations.
Further details on observation-relaxed grouping and OR-GIGPO implementation are provided in Appendix~\ref{app:alg_details}.

\paragraph{Reward Design.}
For a generated rollout $\tau$, we define a composite trajectory-level reward:
\[
R(\tau) = R_{\rm rep}(\tau) + R_{\rm format}(\tau) + R_{\rm correct}(\tau).
\]
The repetition penalty $R_{\rm rep}$ discourages repeated characters, repeated words, and repeated long text spans, and further applies turn-normalized penalties to sentence-level repetition to balance multi-turn tool use. The format reward $R_{\rm format}$ encourages valid trajectories with explicit tags: intermediate tool-use turns should follow \texttt{<think>...</think><action>...</action>}, while the final turn should follow \texttt{<think>...</think><answer>...</answer>}. The correctness reward $R_{\rm correct}$ evaluates the final answer with normalized matching for general spatial QA, covering yes/no, numerical, range, and unordered-list answers. For MapTrace~\citep{panagopoulou2025maptracescalabledatageneration}, we follow its NDTW-based metric and use a continuous reward:
\[
R_{\rm correct}=\min\{1,\max\{0,(d_{\rm high}-d_{\rm NDTW})/(d_{\rm high}-d_{\rm low})\}\}.
\]
Here $d_{\rm NDTW}$ denotes the path distance, with $d_{\rm low}=0.3$ and $d_{\rm high}=0.8$ by default. For other tasks, $R_{\rm correct}$ is binary.
Further details on reward penalties, correctness verification, and the LLM-judge fallback design are provided in Appendix~\ref{app:reward_details}.

\section{Experiments}
\paragraph{Datasets and Metrics.}
Following prior visual reasoning evaluations~\citep{feng2025rewardmap,lai2025minio3scalingreasoningpatterns,lu2025scalingagenticreinforcementlearning,liu2026visionreasonerunifiedreasoningintegratedvisual}, we evaluate \method\ on a broad set of spatial reasoning tasks.
For \textit{in-distribution} evaluation, we select four task families from our training environment with held-out test splits: (1) MapTrace~\citep{panagopoulou2025maptracescalabledatageneration}, (2) ReasonMap~\citep{feng2025can}, (3) ReasonMap-Plus~\citep{feng2025rewardmap}, and (4) Visual Probing tasks~\citep{lai2025minio3scalingreasoningpatterns}.
To assess \textit{out-of-distribution} generalization, we evaluate additional benchmarks not used for training, including four tasks from \citet{wu2026visual} that span distinct spatial reasoning domains: (5) Ball Tracking, (6) Paper Folding, (7) Cube Three-View Reasoning, and (8) Real-world Spatial Reasoning, as well as (9) V*~\citep{vstar}, (10) MapEval~\citep{dihan2025mapevalmapbasedevaluationgeospatial}, and (11) BabyVision~\citep{chen2026babyvisionvisualreasoninglanguage}.
We report accuracy as the unified metric across all tasks.
For MapTrace, we use NDTW to measure path alignment and report threshold-based accuracy, counting a prediction as correct when its distance is below $1.0$. We further discuss how different evaluation criteria affect MapTrace scores in Appendix~\ref{sec:maptrace_details}.
Our training data includes in-distribution trajectories and additional map QA data from MapQA~\citep{chang2022mapqadatasetquestionanswering}.
Detailed information about the training data and evaluation datasets is provided in Appendix~\ref{app:dataset_details}.

\paragraph{Baselines.}
We compare \method\ with three groups of baselines:
(i) \textit{API-based proprietary VLMs}, including GPT-5~\citep{singh2025openai}, Gemini-2.5-Flash~\citep{comanici2025gemini}, and Gemini-2.5-Pro~\citep{comanici2025gemini};
(ii) \textit{open-source VLMs}, including the \texttt{Qwen3-VL} series~\citep{bai2025qwen3vltechnicalreport} with thinking enabled and the InternVL3.5 series~\citep{wang2025internvl35advancingopensourcemultimodal};
and (iii) \textit{tool/reasoning-integrated VLMs}, including VTool-R1 series~\citep{wu2026vtoolr1vlmslearnthink}, R1-OneVision-7B~\citep{yang2025r1onevisionadvancinggeneralizedmultimodal}, and Mini-o3-7B-v1~\citep{lai2025minio3scalingreasoningpatterns}. Non-agentic baselines are evaluated without access to our tool sandbox, while tool-integrated baselines follow their original tool configurations.

\paragraph{Implementation Details.}
We instantiate \method\ using \texttt{Qwen3-VL-Thinking} backbones at 2B, 4B, and 8B scales. We conduct training with Verl-Tool~\citep{jiang2025verltool} and use vLLM~\citep{kwon2023efficient} for inference. For SFT, we train with a learning rate of $1\times10^{-5}$ and a batch size of $32$. For RL, we train with a learning rate of $1\times10^{-6}$, a batch size of $64$, a group size of $N=4$, and $3$ epochs.
Additional implementation details are provided in Appendix~\ref{app:implementation_details}.

\subsection{Main Comparison}
\label{sec:main_comparison}
\begin{table*}[h]
\centering
\footnotesize
\setlength{\tabcolsep}{2.0pt}
\renewcommand\arraystretch{0.93}
\resizebox{1.0\linewidth}{!}{
\begin{tabular}{l @{\hskip2.5pt} | c c c c c c | c | c c c c c c c | c | c}
\toprule
\textbf{Methods ($\downarrow$)}
& \multicolumn{7}{c|}{\textbf{In-Distribution Dataset}}
& \multicolumn{8}{c|}{\textbf{Out-of-Distribution Dataset}}
& \textbf{Overall} \\
\cmidrule(lr){2-8} \cmidrule(lr){9-16} \cmidrule(lr){17-17}
\textbf{Datasets ($\rightarrow$)}
& \makecell{ReasonMap\\Plus}
& ReasonMap
& \makecell{Visual Probing\\Easy}
& \makecell{Visual Probing\\Medium}
& \makecell{Visual Probing\\Hard}
& \makecell{MapTrace}
& Avg.
& MapEval
& BabyVision
& \makecell{Ball\\Tracking}
& \makecell{Paper\\Folding}
& \makecell{Cube 3-View\\Reasoning}
& \makecell{Real-world\\Spatial\\Reasoning}
& V*
& Avg.
& Avg. \\
\midrule
\multicolumn{17}{l}{\textit{Commercial Models}} \\
\midrule
GPT-5
& 89.7 & 53.8 & 62.4 & 21.6 & 25.4 & 82.2 & 55.8
& 72.7 & 22.9 & 25.9 & 11.0 & 32.9 & 38.3 & 70.1 & 39.1
& 46.8 \\
Gemini-2.5-Flash
& 85.4 & 43.2 & 46.1 & 36.0 & 33.6 & 66.4 & 51.7
& 71.0 & 19.5 & 22.7 & 7.5 & 55.0 & 36.7 & 76.9 & 41.3
& 46.2 \\
Gemini-2.5-Pro
& 81.3 & 55.1 & 62.1 & 38.7 & 32.0 & 80.4 & 58.2
& 71.0 & 27.3 & 28.3 & 16.8 & 55.0 & 40.0 & 78.5 & 45.2
& 51.3 \\

\midrule
\multicolumn{17}{l}{\textit{Open-Source VLMs and Tool/Reason-Augmented Baselines}} \\
\midrule
InternVL3.5-8B
& 53.3 & 5.8 & 54.7 & 25.7 & 15.2 & 70.8 & 37.5
& 53.7 & 13.9 & 17.9 & 13.5 & \textbf{51.0} & 28.4 & 71.7 & 35.7
& 36.6 \\
VTool-R1-7B
& 38.3 & 1.5 & 30.5 & 26.4 & 22.6 & 45.2 & 27.4
& 45.7 & 12.3 & 18.0 & \underline{15.2} & 35.6 & 30.5 & 70.6 & 32.5
& 30.2 \\
R1-OneVision-7B
& 52.1 & 8.2 & 39.2 & 21.7 & 17.3 & 41.9 & 30.0
& 45.2 & 14.4 & 15.7 & 12.9 & 33.7 & 28.2 & 53.9 & 29.1
& 29.6 \\
Mini-o3
& 45.9 & 1.6 & \textbf{70.2} & \textbf{52.9} & \underline{41.5} & 30.4 & 40.4
& 46.7 & 11.6 & 16.8 & 12.9 & 39.7 & 28.9 & \textbf{88.2} & 35.0
& 37.5 \\
InternVL3.5-14B
& 51.3 & 4.4 & 58.1 & 12.0 & 19.8 & 76.2 & 36.9
& 54.0 & 14.6 & 16.2 & 11.0 & 46.6 & 26.8 & 69.6 & 34.1
& 35.4 \\
Qwen3-30B-A3B
& 67.8 & 18.2 & 51.7 & 30.2 & 25.4 & 75.0 & 44.7
& 64.7 & 14.9 & 19.5 & 8.5 & 37.7 & 32.8 & 84.2 & 37.4
& 40.8 \\
Qwen3-32B
& 68.8 & 20.6 & 57.4 & 32.0 & 24.5 & 75.5 & 46.4
& \underline{67.2} & \textbf{22.9} & \underline{20.3} & 8.1 & \underline{48.1} & \textbf{38.0} & 85.8 & \textbf{41.4}
& 43.8 \\
Qwen3-235B
& \textbf{76.7} & \underline{30.7} & 58.8 & 32.0 & 27.3 & 72.9 & \underline{49.7}
& \textbf{69.0} & \underline{21.9} & \textbf{20.9} & 6.0 & 38.5 & \underline{34.5} & \underline{86.9} & \underline{39.6}
& \underline{44.3} \\

\midrule
\multicolumn{17}{l}{\textit{Our Method }} \\
\midrule
Qwen3-8B
& 65.7 & 12.3 & 54.6 & 31.3 & 26.6 & 74.1 & 44.1
& 54.0 & 13.0 & 17.8 & 11.0 & 21.2 & 27.7 & 77.5 & 31.7
& 37.4 \\
\rowcolor{gray!15}
\method-8B (ours)
& \underline{72.6}\gain{6.9}
& 27.1\gain{14.8}
& \underline{61.7}\gain{7.1}
& \underline{41.2}\gain{9.9}
& \textbf{44.3}\gain{17.7}
& \underline{77.7}\gain{3.6}
& \textbf{54.1}\gain{10.0}
& 55.5\gain{1.5}
& 15.5\gain{2.5}
& 18.8\gain{1.0}
& \textbf{15.7}\gain{4.7}
& 36.3\gain{15.1}
& 29.1\gain{1.4}
& 82.2\gain{4.7}
& 36.1\gain{4.4}
& \textbf{44.4}\gain{7.0} \\
\rowcolor{gray!8}
\quad\textit{w/o RL}
& 66.2
& 21.1
& 51.0
& 37.3
& 32.0
& 74.2
& 46.9
& 54.8
& 12.4
& 18.1
& 14.6
& 39.0
& 29.6
& 74.4
& 34.7
& 40.4 \\

\rowcolor{gray!8}
\quad\textit{w/o SFT}
& 64.2
& \textbf{45.1}
& 19.1
& 18.6
& 26.4
& 71.1
& 40.7
& 49.5
& 7.2
& 12.4
& 6.0
& 29.1
& 25.0
& 59.2
& 26.9
& 33.3 \\

\rowcolor{gray!8}
\quad\textit{w/o Tool}
& 72.4
& 13.7
& 36.1
& 21.2
& 19.8
& \textbf{82.4}
& 40.9
& 59.0
& 13.9
& 19.6
& 15.0
& 30.4
& 28.3
& 76.4
& 34.7
& 37.6 \\
\bottomrule
\end{tabular}%
}
\caption{
\textbf{\method\ closes much of the gap to larger proprietary and open-source models.}
Scores are accuracy percentages, green gains are percentage points, and Qwen3 denotes Qwen3-VL-Thinking models.
\method-8B substantially improves over strong open-source and tool/reasoning-augmented baselines, achieving the best overall average among non-commercial models and performance comparable to GPT-5 and Qwen3-235B.
We also report ablations: \textit{w/o RL} uses SFT alone, \textit{w/o SFT} trains directly with RL from the base model, and \textit{w/o Tool} removes tool access during training and inference.
}
\label{tab:main_results}
\end{table*}

\textbf{\method-8B\ achieves strong overall performance across 13 benchmarks.}
As reported in Table~\ref{tab:main_results}, \method-8B consistently improves over the Qwen3-8B backbone by $10.0$ percentage points on in-distribution tasks and $4.4$ percentage points on out-of-distribution tasks, and it exceeds prior state-of-the-art baselines at a comparable scale by $6.9$--$14.8$ percentage points across 13 benchmarks from 8 datasets. On average, it further approaches larger open-weight and proprietary models such as Qwen3-VL-235B-A22B-Thinking and GPT-5.
Across individual domains, including map-based reasoning, route tracing, visual probing, real-world spatial understanding, dynamic object tracking, object transformation, multi-view 3D reasoning, visual search, geospatial understanding, and language-free visual reasoning, \method-8B\ outperforms most similarly sized baselines and achieves performance comparable to much larger open-source models. These results indicate that effective visual tool-use training can provide gains beyond backbone scaling alone.

\textbf{SFT establishes tool competence, while the subsequent RL stage adds substantial gains.}
In Table~\ref{tab:main_results}, the backbone-to-SFT comparison (the \textit{w/o RL} checkpoint) increases the six-ID average from $44.1\%$ to $46.9\%$ (+2.8), the OOD average from $31.7\%$ to $34.7\%$ (+3.0), and the overall average from $37.4\%$ to $40.4\%$ (+3.0).
Adding the full RL stage then increases the ID average to $54.1\%$ (+7.2), the OOD average to $36.1\%$ (+1.4), and the overall average to $44.4\%$ (+4.0).
Thus, the OOD decrease when RL is removed is $1.4$ percentage points; $4.0$ percentage points is the overall decrease.
The \textit{w/o SFT} setting applies RL directly to the base model and answers whether tool use can be bootstrapped without supervised initialization; its $40.7\%$ six-ID average is not an additive estimate of SFT's contribution.
Within the matched RL optimizer comparison in Table~\ref{tab:ablation}, OR-GIGPO is the strongest tested optimizer, reaching $54.1\%$ versus $51.3\%$ for exact GIGPO, $51.7\%$ for GRPO, and $43.0\%$ for DAPO.

\textbf{The \method\ tool sandbox and trajectory synthesis are essential for spatial reasoning.}
The \textit{w/o Tool} ablation in Table~\ref{tab:main_results} shows an imbalanced performance pattern, suggesting that training without tool access can lead to catastrophic forgetting in certain spatial domains and fails to support robust performance across diverse tasks.
Meanwhile, the \textit{w/o SFT} variant performs substantially worse than the full \method, showing that synthesized tool-use trajectories are critical for teaching reliable tool invocation before RL. 
The inference-time \textit{Tool-use ablation} in Table~\ref{tab:ablation} further confirms the complementary roles of perception and interaction tools. These results support our toolset design and show that \method\ benefits from both spatially targeted tools and supervised trajectory synthesis.

\subsection{Ablation Studies}
\label{sec:ablation}

\begin{table}[h]
\centering
\footnotesize
\setlength{\tabcolsep}{3pt}
\renewcommand\arraystretch{1.1}
\resizebox{\linewidth}{!}{%
\begin{tabular}{l | c c c c c c | c}
\toprule
\makecell{\textbf{Datasets}\textbf{($\rightarrow$)}\\\textbf{Model ($\downarrow$)}}
& \makecell{\textbf{ReasonMap}\\\textbf{Plus}}
& \textbf{ReasonMap}
& \makecell{\textbf{Visual Probing}\\\textbf{Easy}}
& \makecell{\textbf{Visual Probing}\\\textbf{Medium}}
& \makecell{\textbf{Visual Probing}\\\textbf{Hard}}
& \makecell{\textbf{MapTrace}}
& \textbf{Avg.} \\
\midrule
\multicolumn{8}{l}{\textit{RL algorithm ablation}} \\
\midrule
Qwen3-8B
& 65.7 & 12.3 & 54.6 & 31.3 & 26.6 & 74.1 & 44.1 \\
\rowcolor{gray!15}
\method-8B (ours)
& & & & & & & \\
\rowcolor{gray!15}
\quad \textit{w/ OR-GIGPO}
& \textbf{72.6} & \textbf{27.1} & \textbf{61.7} & 41.2 & \textbf{44.3} & \textbf{77.7} & \textbf{54.1} \\
\rowcolor{gray!8}
\quad \textit{w/ GIGPO}
& 70.6 & 21.8 & \textbf{61.7} & 40.3 & 42.5 & 70.6 & 51.3 \\
\rowcolor{gray!8}
\quad \textit{w/ GRPO}
& 70.0 & 24.3 & 58.8 & \textbf{41.5} & 40.6 & 75.5 & 51.7 \\
\rowcolor{gray!8}
\quad \textit{w/ DAPO}
& 69.2 & 20.8 & 43.2 & 32.4 & 27.3 & 65.3 & 43.0 \\
\midrule
\multicolumn{8}{l}{\textit{Model-size ablation}} \\
\midrule
Qwen3-2B
& 53.1 & 8.8 & 42.8 & 21.7 & 23.0 & 66.2 & 35.9 \\
\rowcolor{gray!15}
\method-2B
& \textbf{58.2} & \textbf{16.7} & \textbf{43.0} & \textbf{26.0} & 19.8 & \textbf{72.2} & \textbf{39.3} \\
\midrule
Qwen3-4B
& 64.2 & 13.7 & 46.1 & 25.3 & 26.4 & 71.6 & 41.2 \\
\rowcolor{gray!15}
\method-4B
& \textbf{67.1} & \textbf{21.4} & \textbf{58.1} & \textbf{36.1} & \textbf{35.8} & \textbf{74.5} & \textbf{48.8} \\
\midrule
\multicolumn{8}{l}{\textit{Tool-use ablation}} \\
\midrule
Qwen3-8B Baseline
& 65.7 & 12.3 & 54.6 & 31.3 & 26.6 & 74.1 & 44.1 \\
\rowcolor{gray!15}
\method-8B (ours)
& \textbf{72.6} & \textbf{27.1} & \textbf{61.7} & \textbf{41.2} & \textbf{44.3} & 77.7 & \textbf{54.1} \\
\rowcolor{gray!8}
\quad \textit{No Tool}
& 67.6 & 26.6 & 46.9 & 23.9 & 25.5 & 75.9 & 44.4 \\
\rowcolor{gray!8}
\quad \textit{w/ Previous Tool Set}
& 65.8 & 22.1 & 58.9 & 36.2 & 25.5 & 77.6 & 47.6 \\
\rowcolor{gray!8}
\quad \textit{w/o Perception Tools}
& 46.0 & 11.2 & 57.5 & 34.7 & 34.9 & 76.4 & 43.5 \\
\rowcolor{gray!8}
\quad \textit{w/o Interaction Tools}
& 69.4 & 25.0 & 53.2 & 26.5 & 32.1 & \textbf{78.5} & 47.4 \\
\bottomrule
\end{tabular}%
}%
\caption{
\textbf{OR-GIGPO, model scale, and tool access all affect performance.}
OR-GIGPO gives the best 8B result among RL algorithms, and the same recipe improves the 2B and 4B backbones.
The full tool sandbox also outperforms no-tool, previous-tool, and tool-family removal settings, showing that perception and interaction tools are complementary. All scores are accuracy percentages.
}
\label{tab:ablation}
\end{table}

\textbf{Observation-relaxed credit assignment improves multi-step tool learning.} Table~\ref{tab:ablation} compares OR-GIGPO with original GIGPO, GRPO, and DAPO on in-distribution benchmarks.
OR-GIGPO achieves the best average accuracy, outperforming GIGPO by $2.8$, GRPO by $2.4$, and DAPO by $11.1$ percentage points, which supports the value of observation-relaxed step-level advantages for fine-grained credit assignment in multi-step tool use.
The \textit{Average Step Adv Ratio} measures the fraction of trajectories with any nonzero step advantage: $9.3\%$ for GIGPO and $62.3\%$ for OR-GIGPO.
This indicates that, in the setting of visual tool agents, original GIGPO struggles to match exactly identical observations and therefore largely degrades to GRPO-level behavior, while OR-GIGPO constructs relaxed step-level outcome groups that enable fine-grained advantage assignment.
The similarity threshold must remain selective: lowering it from $\delta=0.9$ to $\delta=0.5$ further raises the step advantage ratio to $68.5\%$ but reduces the average accuracy to $48.9\%$, as dissimilar observations are merged into the same outcome group.
DAPO performs worse than the baseline, suggesting that its algorithmic design may not be well suited to our visual tool-use setting, where early-stage tool-calling ability is still unstable.

\textbf{\method's training recipe is effective beyond the 8B model.} Table~\ref{tab:ablation} also shows that the proposed training recipe transfers to smaller backbones.
Compared with their Qwen3 baselines, \method-2B improves from $35.9\%$ to $39.3\%$, and \method-4B improves from $41.2\%$ to $48.8\%$ on average.
This indicates the generality of \method\ across model scales.

\textbf{Both perception and interaction tools are needed for robust spatial reasoning.} Table~\ref{tab:ablation} evaluates \method-8B under different inference-time tool conditions.
Removing all tools reduces the average score to $44.4\%$, showing that tool access is necessary for robust performance across diverse spatial reasoning domains.
Using only the previous tool set reaches $47.6\%$, clearly below the full sandbox; this previous set contains six tools adopted from existing work: \texttt{image\_crop}, \texttt{bounding\_box}, \texttt{image\_highlight}, \texttt{text\_ocr}, \texttt{auto\_segment}, and \texttt{grounding\_dino}.
This gap demonstrates that our \method\ tool design is more effective than generic visual utilities for spatial reasoning.
Removing either perception or interaction tools also degrades performance, confirming their complementary roles in evidence acquisition, local verification, and spatial planning.

\textbf{Additional ablations and analyses.} We provide further details and studies in the appendix.
Appendix~\ref{app:interaction_cap} reports the interaction-cap diagnostic, followed by detailed tool definitions (Appendix~\ref{app:tool_sandbox}) and the cost of thought diversification (Appendix~\ref{app:diversification_cost}).
Appendix~\ref{sec:maptrace_details} examines the MapTrace evaluation criterion.
Appendix~\ref{app:additional_experiments} studies trajectory-teacher quality (Appendix~\ref{app:teacher_quality}) and the rollout group size of OR-GIGPO (Appendix~\ref{app:group_size}).
Appendix~\ref{app:tool_analysis} then presents fine-grained tool-use analyses: tool-call distribution (Appendix~\ref{app:tool_distribution}), zero-shot transfer of the \method\ sandbox to prior agents (Appendix~\ref{app:interface_transfer}), map-oriented-tool removal and map/non-map balance (Appendix~\ref{app:map_tools}), and tool effectiveness on Cube Three-View (Appendix~\ref{app:cube}).

\subsection{Qualitative Analysis}

\begin{figure}[h]
\centering
\includegraphics[width=\linewidth]{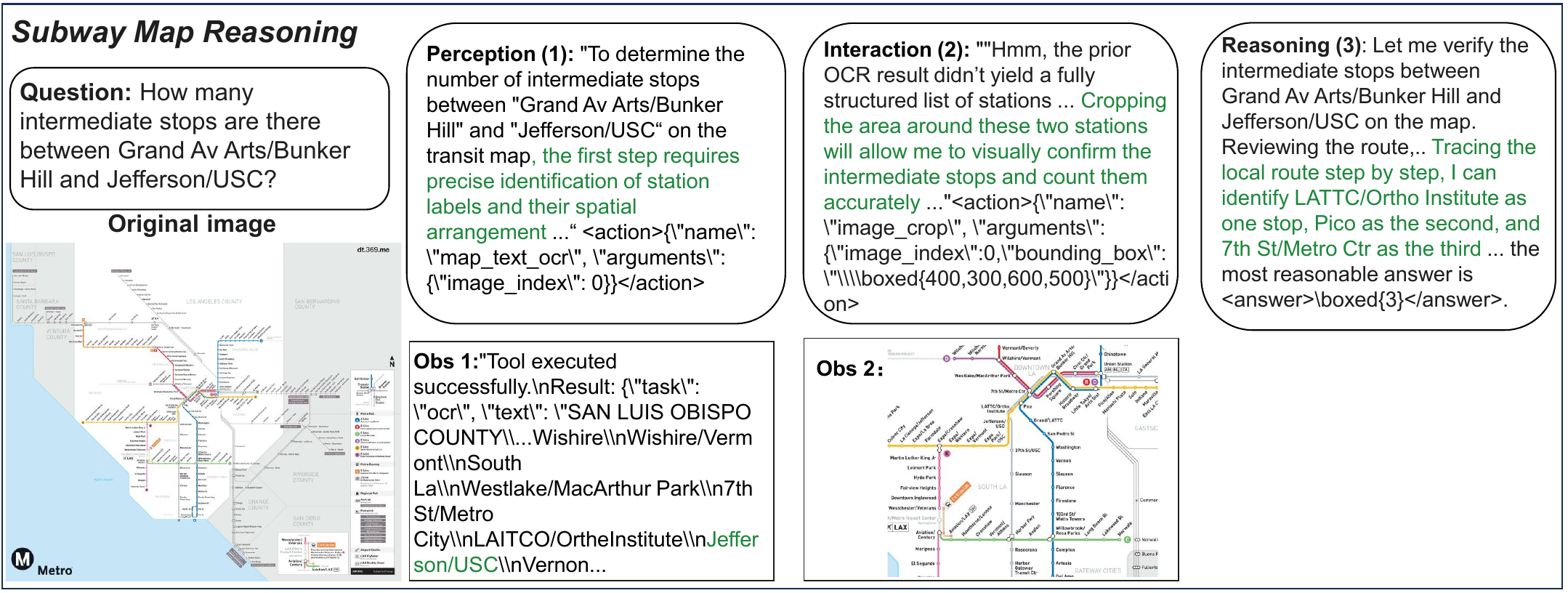}
\caption{
\textbf{\method\ grounds route tracing with visual tools.}
In this qualitative example, \method\ uses perception tools to identify station labels, interaction tools to verify the relevant region, and accumulated observations to trace the route and answer correctly.
}
\label{fig:qualitative_result}
\end{figure}

Figure~\ref{fig:qualitative_result} illustrates how \method\ combines perception and interaction for subway-map reasoning.
Counting intermediate stops between Grand Av Arts/Bunker Hill and Jefferson/USC requires recovering the connecting station sequence and excluding the endpoints.
Small labels and intersecting lines make this difficult to verify from the full map alone.

\textbf{Perception identifies relevant entities; interaction verifies their spatial relationships.}
\method\ first invokes \texttt{map\_text\_ocr} to extract station labels.
The observation contains relevant names but lacks a sufficiently structured sequence for reliable counting.
The agent therefore invokes \texttt{image\_crop} to inspect the queried region and verify the local route.
It identifies LATTC/Ortho Institute, Pico, and 7th St/Metro Ctr as the three intermediate stops and returns $3$.
The tools play complementary roles: perception narrows the search, while interaction resolves local spatial uncertainty.
The answer follows from reasoning over both observations rather than accepting the initial OCR output as a complete solution.
More examples appear in Appendix~\ref{app:case_studies}.

\section{Conclusion}

We presented \methodfull, a tool-augmented visual agent framework for general-purpose spatial reasoning. \method\ integrates perception and interaction tools into a unified Perceive--Interact--Reason process, enabling agents to acquire global spatial evidence, perform fine-grained local verification, and reason over accumulated observations. To train such agents, we introduced supervised tool-use trajectory synthesis and OR-GIGPO, an observation-relaxed policy optimization method for multi-step visual tool use. Experimental results show that \method\ consistently improves spatial reasoning performance, approaches much larger proprietary and open-source models, and benefits from both the designed tool sandbox and the two-stage training recipe. These results suggest that trained visual tool use is a promising direction for building more capable spatial reasoning agents. We discuss limitations and future work in Appendix~\ref{app:limit_future}.

\paragraph{Broader impact.}
\method\ co-designs a heterogeneous spatial tool sandbox, executable trajectory supervision, and observation-relaxed credit assignment, making multi-tool spatial reasoning agents trainable.
Our findings suggest that tool access alone does not yield tool competence, that executable and verified trajectories are needed to establish it, and that credit assignment should tolerate the surface-form variation of tool observations.
We therefore expect future tool-augmented multimodal agents to benefit from designing the tool interface, trajectory supervision, and credit assignment jointly rather than in isolation.

Such agents may support visual assistance, map understanding, robotics, and navigation, but incorrect tool-grounded reasoning may mislead safety-sensitive decisions and the framework could be misused for surveillance; responsible deployment should include task-specific evaluation and human oversight.

\begin{ack}
This work was supported by research funds from Xiong'an AI Institute.

Beyond the primary affiliations listed above, the authors declare
no financial relationships during the 36 months prior to submission
that could reasonably be perceived to have influenced this work.
\end{ack}

\bibliographystyle{plainnat}
\bibliography{references}

\clearpage
\appendix

\addtocontents{toc}{\protect\setcounter{tocdepth}{2}}
\begingroup
\hypersetup{hidelinks}
\renewcommand{\contentsname}{Appendix Contents}
\tableofcontents
\endgroup
\clearpage

\section{Limitations and Future Work}
\label{app:limit_future}

Although \method\ shows consistent gains across model sizes, our main experiments focus on the 8B model due to computational constraints. We leave larger-backbone validation and a more detailed study of scaling with trajectory data size and tool-use diversity to future work. In addition, we restrict inference to at most 10 interaction turns. This setting covers the tasks studied in this paper, but longer-horizon spatial reasoning may require stronger memory, better exploration, and more stable credit assignment over extended tool-use trajectories.

Future work will extend \method\ to broader spatial reasoning domains, including realistic 3D reasoning, maze navigation, Sokoban-like planning, and embodied or environment-centric tasks that require stronger spatial state tracking. Building larger and more diverse trajectory datasets is also important for exploring the upper bound of \method.

\section{Implementation Details}
\label{app:implementation_details}

\subsection{Training and Inference}

\paragraph{SFT training parameters.}
We first perform supervised fine-tuning to initialize the model with basic tool-use and spatial reasoning abilities. 
Table~\ref{tab:sft_training_parameters} summarizes the main SFT hyperparameters, including the training schedule, optimization settings, and cutoff length.

\begin{table}[ht]
    \centering
    \footnotesize
    \begin{tabular}{lc}
        \toprule
        \textbf{Parameters} & \textbf{SFT Training} \\
        \midrule
        epochs & 1.0 \\
        global batch size & 32 \\
        learning rate & $1.0 \times 10^{-5}$ \\
        scheduler & cosine \\
        warmup ratio & 0.1 \\
        weight decay & 0.01 \\
        cutoff length & 65,536 \\
        \bottomrule
    \end{tabular}
    \caption{
    \textbf{SFT Parameters.}
    We report the main optimization and cutoff-length settings used during supervised fine-tuning.
    }
    \label{tab:sft_training_parameters}
\end{table}

\paragraph{RL training parameters.}
After SFT, we further optimize the model with reinforcement learning to improve reliable tool use and multi-step spatial reasoning. 
Table~\ref{tab:rl_training_parameters} reports the shared RL training and rollout settings, together with the algorithm-specific parameters used by OR-GIGPO and DAPO.

\begin{table}[ht]
    \centering
    \footnotesize
    \begin{tabular}{lc}
        \toprule
        \textbf{Parameters} & \textbf{RL Training} \\
        \midrule
        rollout samples per prompt, $N$ & 4 \\
        train batch size & 64 \\
        PPO mini-batch size & 256 \\
        PPO micro-batch size / GPU & 4 \\
        log-prob micro-batch size / GPU & 16 \\
        actor learning rate & $1.0 \times 10^{-6}$ \\
        learning rate scheduler & cosine \\
        warmup ratio & 0.05 \\
        total epochs & 3 \\
        max prompt length & 16,384 \\
        max response length & 32,768 \\
        max action length & 4,096 \\
        max observation length & 8,192 \\
        max turns, $T$ & 11 \\
        sampling temperature & 1.0 \\
        top\_p & 1.0 \\
        top\_k & -1 \\
        rollout backend & vLLM \\
        standard clipping coefficient, $\epsilon_{\rm clip}$ & 0.2 \\
        \midrule
        \multicolumn{2}{c}{\textit{OR-GIGPO-specific parameters}} \\
        \midrule
        episode-step advantage weight, $\omega$ & 1.0 \\
        discount factor, $\gamma$ & 0.99 \\
        advantage normalization constant, $\epsilon_{\rm adv}$ & $1.0 \times 10^{-6}$ \\
        observation similarity threshold, $\delta$ & 0.9 \\
        \midrule
        \multicolumn{2}{c}{\textit{DAPO-specific parameters}} \\
        \midrule
        lower clipping coefficient, $\epsilon_{\rm low}$ & 0.2 \\
        upper clipping coefficient, $\epsilon_{\rm high}$ & 0.28 \\
        total training steps & 300 \\
        \bottomrule
    \end{tabular}
    \caption{
    \textbf{RL Parameters.}
    We report the rollout configuration, optimization settings, interaction limits, and algorithm-specific parameters for OR-GIGPO and DAPO.
    }
    \label{tab:rl_training_parameters}
\end{table}

\paragraph{Inference parameters.}
Table~\ref{tab:model_inference_parameters} summarizes the inference configuration used to evaluate \method.

\begin{table}[ht]
    \centering
    \footnotesize
    \begin{tabular}{lc}
        \toprule
        \textbf{Parameters} & \textbf{Inference} \\
        \midrule
        max response length & 32,768 \\
        max turns & 11 \\
        temperature & 0.0 \\
        inference backend & vLLM \\
        \bottomrule
    \end{tabular}
    \caption{
    \textbf{Inference Parameters.}
    Parameters used for \method\ evaluation.
    }
    \label{tab:model_inference_parameters}
\end{table}

\begin{figure}[ht]
    \centering
    \includegraphics[width=0.6\linewidth]{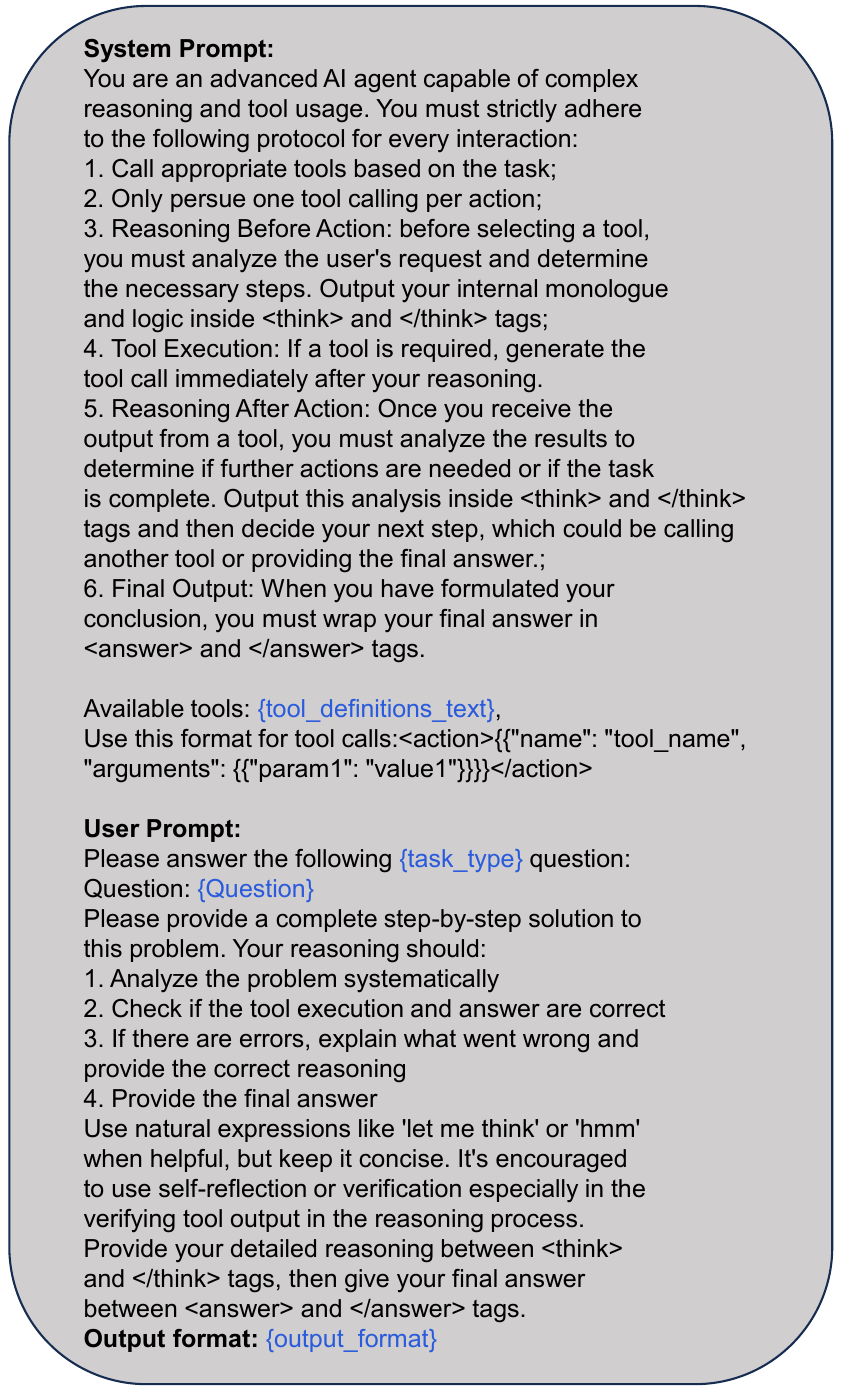}
    \caption{
    \textbf{The inference prompt enforces structured tool use and answer formatting.}
    It defines the tool-use protocol, including tool selection, one tool call per action, pre- and post-action reasoning, observation verification, and the required final-answer format.
    }
    \label{fig:inference_prompt}
\end{figure}

\paragraph{Inference prompts.}
We use the inference prompt below to evaluate \method. 
As shown in Figure~\ref{fig:inference_prompt}, the prompt defines the available tools, enforces one tool call per action, and requires the model to reason before and after each tool execution. 
The model is instructed to place intermediate reasoning inside \texttt{<think>} and \texttt{</think>} tags, issue tool calls using the prescribed \texttt{<action>} format, and provide the final answer inside \texttt{<answer>} and \texttt{</answer>} tags. 
This protocol encourages the agent to verify tool observations before producing the final response.

\paragraph{Interaction cap.}\label{app:interaction_cap}
The maximum setting of $T=11$ consists of at most ten tool-interaction rounds followed by one final-answer round.  Profiling the SFT-only checkpoint on a 10\% sample of RL-training prompts found that 1.16\% of trajectories reached the tenth interaction and 0.81\% attempted to continue beyond it and were truncated; 99.19\% were not truncated.

\subsection{Evaluation Protocol}
\label{app:evaluation_details}
\begin{figure}[ht]
    \centering
    \includegraphics[width=0.6\linewidth]{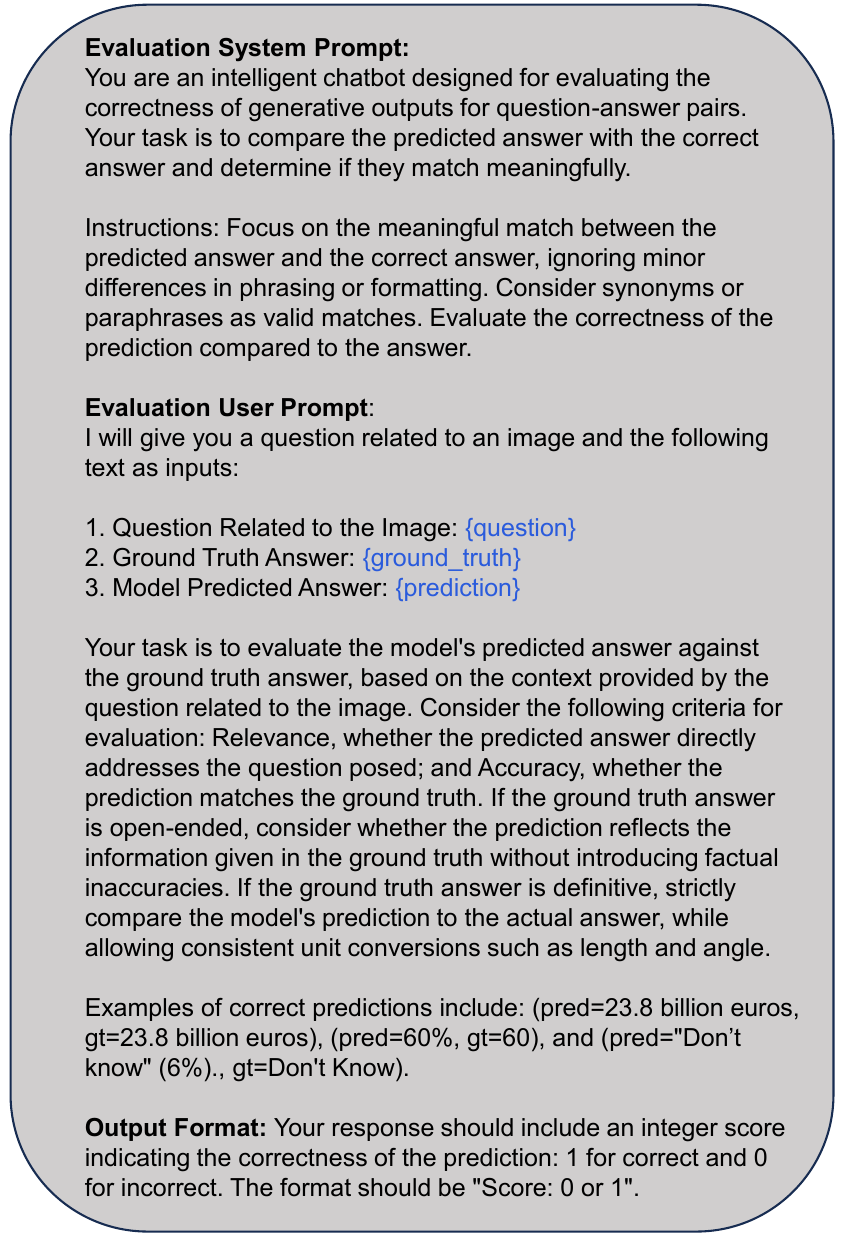}
    \caption{
    \textbf{The evaluation prompt checks semantic correctness beyond exact matching.}
    It asks the LLM judge to compare the model prediction with the ground-truth answer for an image-related question and output a binary correctness score. 
    The prompt emphasizes semantic matching, relevance, accuracy, and tolerance to minor formatting or phrasing differences.
    }
    \label{fig:eval_prompt}
\end{figure}

\paragraph{Evaluation protocol.}
We adopt a two-stage evaluation protocol for answer-based spatial reasoning tasks. 
We first apply rule-based matching to compare model predictions with ground-truth answers, which handles exact matches, normalized formatting, and simple equivalent forms. 
If the rule-based matcher does not mark the prediction as correct, we further use an LLM judge (\texttt{GPT-5-Mini}) to evaluate whether the prediction meaningfully matches the ground truth. 
The judge focuses on semantic correctness and ignores minor differences in phrasing or formatting, while allowing equivalent expressions such as unit conversions when they are consistent with the ground truth.

For MapTrace, where the output is a coordinate path rather than a textual answer, we directly evaluate the predicted trajectory against the ground-truth trajectory using normalized dynamic time warping (NDTW). 
Both paths are represented as normalized coordinates in $[0,1]$, and the cumulative DTW cost is computed with Euclidean distance between matched points. 
We regard a prediction as correct if its NDTW score is below the threshold of 1.0. 
Lower NDTW indicates better alignment between the predicted and ground-truth paths.

\paragraph{Evaluation prompts.}
For predictions that cannot be verified by rule-based matching, we use the evaluation prompt shown in Figure~\ref{fig:eval_prompt}. 
The prompt provides the image-related question, the ground-truth answer, and the model prediction, and asks the judge to output a binary correctness score. 
The judge is instructed to consider both relevance and accuracy, allowing paraphrases and formatting differences while rejecting predictions that introduce factual errors.

\section{Method Details}
\label{app:method}

\subsection{Tool Sandbox}
\label{app:tool_sandbox}

We organize the tool sandbox into two categories: perception tools and interaction tools. 
Perception tools transform images into structured observations, including text, localized text boxes, object detections, segmentation proposals, counts, and presence signals, powered by three lightweight tool agents: PaddleOCR-VL-1.5~\citep{cui2026paddleocrvl15multitask09bvlm}, Grounding DINO~\citep{liu2024groundingdinomarryingdino}, and SAM3~\citep{carion2026sam3segmentconcepts}. 
The OCR agent is served with vLLM using a maximum context length of $8192$ and maximum generation length of $4096$; the detection agent uses GroundingDINO base with box/text thresholds of $0.25$ and NMS threshold of $0.8$; and the segmentation agent uses SAM3.1 with score threshold $0.25$ and at most $20$ returned proposals. 
Interaction tools perform deterministic image operations, such as cropping, labeling, drawing, and highlighting, to help the agent inspect and verify spatial evidence during multi-step reasoning.

All tools use \texttt{image\_index} to specify the input image. 
For tools requiring spatial coordinates, we use normalized coordinates in a $1000 \times 1000$ image space. 
Bounding boxes follow the format \texttt{\textbackslash boxed\{x1,y1,x2,y2\}}, where $(x1,y1)$ and $(x2,y2)$ denote the top-left and bottom-right corners. 
Table~\ref{tab:tool_sandbox_overview} summarizes the full tool list, including each tool's execution backend, output type, and main parameters.

\begin{table}[ht]
    \centering
    \footnotesize
    \setlength{\tabcolsep}{4pt}
    \renewcommand{\arraystretch}{1.08}
    \begin{tabular}{l l l p{0.34\linewidth}}
        \toprule
        \textbf{Tool} & \textbf{Driven by} & \textbf{Output} & \textbf{Main parameters} \\
        \midrule
        \multicolumn{4}{l}{\textit{Visual perception tools}} \\
        \midrule
        \texttt{text\_ocr} & PaddleOCR-VL-1.5 0.9B & Text & \texttt{image\_index} \\
        \texttt{text\_spotting} & PaddleOCR-VL-1.5 0.9B & Text & \texttt{image\_index} \\
        \texttt{map\_text\_ocr} & PaddleOCR-VL-1.5 0.9B & Text & \texttt{image\_index} \\
        \texttt{formula\_ocr} & PaddleOCR-VL-1.5 0.9B & Text & \texttt{image\_index} \\
        \texttt{table\_ocr} & PaddleOCR-VL-1.5 0.9B & Text & \texttt{image\_index} \\
        \texttt{grounding\_dino} & GroundingDINO base & Text & \texttt{image\_index}, \texttt{question} \\
        \texttt{auto\_segment} & SAM3.1 & Text & \texttt{image\_index} \\
        \texttt{bbox\_segment} & SAM3.1 & Text & \texttt{image\_index}, \texttt{bounding\_box} \\
        \texttt{text\_segment} & SAM3.1 & Text & \texttt{image\_index}, \texttt{text\_prompt} \\
        \texttt{exemplar\_segment} & SAM3.1 & Text & \texttt{image\_index}, \texttt{bounding\_box} \\
        \texttt{concept\_count} & SAM3.1 & Text & \texttt{image\_index}, \texttt{text\_prompt} \\
        \texttt{presence\_check} & SAM3.1 & Text & \texttt{image\_index}, \texttt{text\_prompt} \\

        \midrule
        \multicolumn{4}{l}{\textit{Visual interaction tools}} \\
        \midrule
        \texttt{image\_crop} & Function tool & Image & \texttt{image\_index}, \texttt{bounding\_box} \\
        \texttt{image\_label} & Function tool & Image & \texttt{image\_index}, \texttt{text}, \texttt{position} \\
        \texttt{draw\_line} & Function tool & Image & \texttt{image\_index}, \texttt{coordinates} \\
        \texttt{draw\_path} & Function tool & Image & \texttt{image\_index}, \texttt{points} \\
        \texttt{bounding\_box} & Function tool & Image & \texttt{image\_index}, \texttt{bounding\_box} \\
        \texttt{image\_highlight} & Function tool & Image & \texttt{image\_index}, \texttt{bounding\_box} \\
        \bottomrule
    \end{tabular}
    \caption{
\textbf{Complete tool list in the \method\ Tool Sandbox.}
We list all perception and interaction tools, together with their execution backends, output types, and main input parameters.
}
    \label{tab:tool_sandbox_overview}
\end{table}

\paragraph{Perception tools.}
The perception tools extract structured visual evidence from the image.

\begin{itemize}
    \item \textbf{\texttt{text\_ocr}} performs general OCR and extracts visible text from images, such as document text, signs, labels, and natural scene text.

    \item \textbf{\texttt{text\_spotting}} detects and recognizes text with localization, returning both text content and bounding boxes. 
    It is useful when the spatial position of text matters, such as station names, road labels, or map annotations.

    \item \textbf{\texttt{map\_text\_ocr}} extracts map-related text, including place names, road names, station names, and landmarks. 
    It filters common map OCR noise such as pure numbers, scale markers, and isolated symbols.

    \item \textbf{\texttt{formula\_ocr}} recognizes mathematical expressions and equations from images and converts them into textual or \LaTeX-style outputs.

    \item \textbf{\texttt{table\_ocr}} recognizes table structures and extracts rows, columns, and cell contents from structured images.

    \item \textbf{\texttt{grounding\_dino}} performs open-vocabulary object detection from text descriptions. 
    It returns detected objects with bounding boxes, confidence scores, and labels, which helps locate landmarks, icons, objects, or semantic regions.

    \item \textbf{\texttt{auto\_segment}} automatically segments objects in the full image and returns proposals with bounding boxes, confidence scores, areas, and centroids.

    \item \textbf{\texttt{bbox\_segment}} performs region-constrained segmentation inside a specified bounding box, enabling focused analysis or refinement of a target area.

    \item \textbf{\texttt{text\_segment}} segments objects described by a natural-language prompt, supporting open-vocabulary object localization and isolation.

    \item \textbf{\texttt{exemplar\_segment}} uses a bounding box as a visual exemplar and finds visually similar objects in the image, which is useful for repeated patterns or ``find more like this'' reasoning.

    \item \textbf{\texttt{concept\_count}} counts objects matching a text prompt and returns both the count and instance locations.

    \item \textbf{\texttt{presence\_check}} checks whether a described concept is present in the image and returns a presence flag, confidence score, and count.
\end{itemize}

\paragraph{Interaction tools.}
The interaction tools return modified images and support active visual inspection and verification.

\begin{itemize}
    \item \textbf{\texttt{image\_crop}} crops a specified region for fine-grained inspection, such as verifying small text, landmarks, or local route details.

    \item \textbf{\texttt{image\_label}} adds a text label at a specified position, helping annotate landmarks, objects, or intermediate reasoning results.

    \item \textbf{\texttt{draw\_line}} draws a single line between two points, which is useful for marking road segments, boundaries, crossings, or direct spatial relations.

    \item \textbf{\texttt{draw\_path}} draws a connected path through multiple waypoints, supporting route tracing, maze solving, and multi-step path verification.

    \item \textbf{\texttt{bounding\_box}} draws a bounding box around a target region, providing explicit localization of objects, landmarks, or regions of interest.

    \item \textbf{\texttt{image\_highlight}} highlights a specified region with a translucent overlay, helping focus attention on relevant areas while preserving surrounding context.
\end{itemize}

\subsection{Trajectory Synthesis}
\label{app:Trajectory_Synthesis}

\begin{figure}[h]
    \centering
    \includegraphics[width=1.0\linewidth]{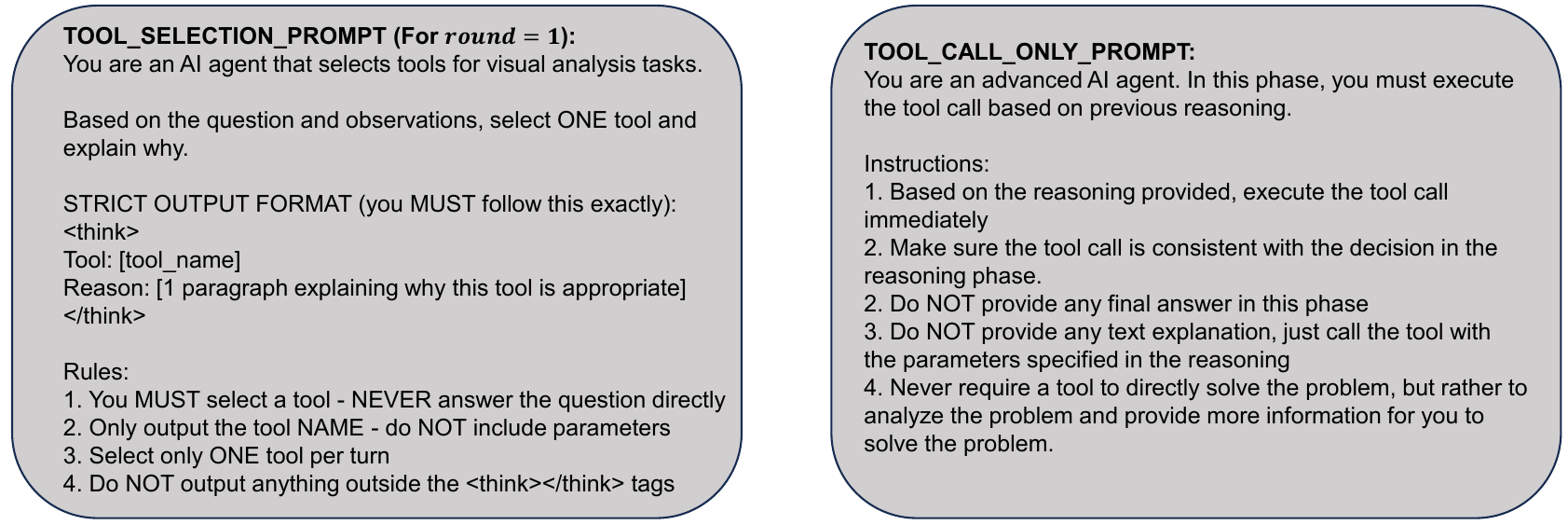}
    \caption{
    \textbf{Round-1 prompts force exploration before answering.}
    At round $1$, \texttt{TOOL\_SELECTION\_PROMPT} forces the model to select exactly one tool and explain the reason inside \texttt{<think>} tags, while \texttt{TOOL\_CALL\_ONLY\_PROMPT} executes the selected tool with concrete parameters and forbids final answers.
    }
    \label{fig:sys_prompt1}
\end{figure}

\paragraph{Data synthesis procedure.}
We synthesize tool-use trajectories with an iterative explore-and-exploit procedure. 
For each example, we initialize the task with the image, question, ground-truth answer, dataset-specific metadata, and the available tool sandbox. 
We set the number of sampled trajectories per round to $n_{\rm sample}=3$ and the maximum number of rounds to $T=11$, including at most $10$ tool-use rounds and one final-answer round.
Rather than asking the model to answer directly, we decompose trajectory generation into tool selection, tool execution, observation gathering, and final answering.

At round $1$, the model is forced to explore by selecting one tool for visual analysis and then executing the corresponding tool call. 
For rounds $2$ to $T-1$, the model exploits accumulated observations to decide whether the current evidence is sufficient for answering or whether another tool call is needed. 
If the model produces an answer before the final round, we validate it before committing it to the trajectory. 
Correct answers are saved as valid trajectories, while rejected answers are removed from the conversation history and trigger a reflection step that forces the model to gather additional evidence with a new or more informative tool. 
At round $T$, if no valid trajectory has been found, the model is prompted to reflect on all observations and provide a final answer.

The notation $y_t^{(i)}=y^\ast$ in the main text denotes correctness under the task-specific verifier. We validate synthesized trajectories using the same task-specific criteria as our evaluation protocol in Appendix~\ref{app:evaluation_details}, including rule-based and LLM-judge checks for answer-based tasks and NDTW-based verification for MapTrace. 
This process yields executable trajectories with explicit tool choices, grounded observations, self-reflection after rejected answers, and verified final responses.

\begin{figure}[ht]
    \centering
    \includegraphics[width=1.0\linewidth]{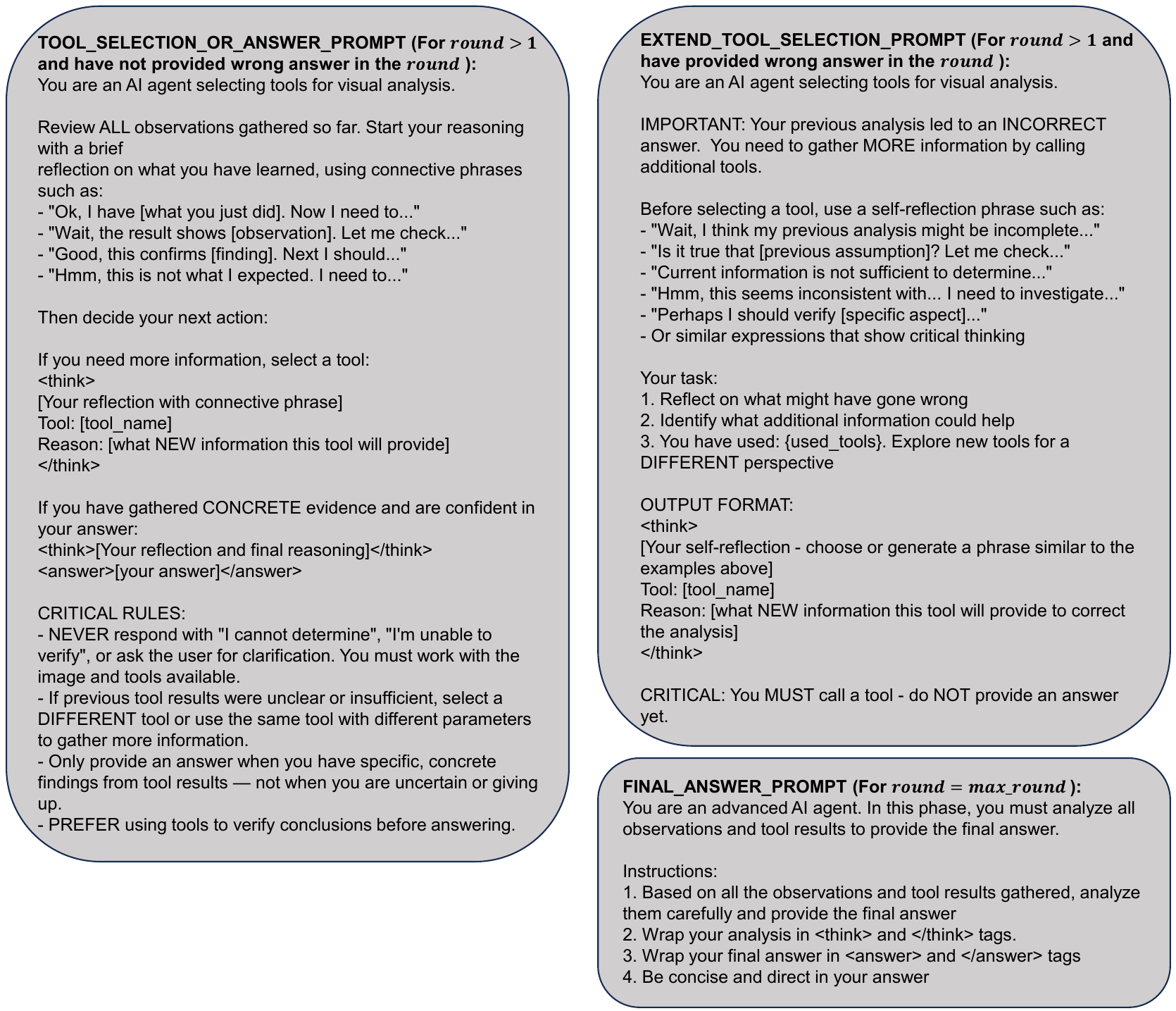}
    \caption{
    \textbf{Later-round prompts balance further tool use with final answering.}
    For rounds $>1$, \texttt{TOOL\_SELECTION\_OR\_ANSWER\_PROMPT} lets the model choose between further tool use and final answering; after a rejected answer, \texttt{EXTEND\_TOOL\_SELECTION\_PROMPT} forces self-reflection and additional tool use; at the maximum round, \texttt{FINAL\_ANSWER\_PROMPT} requires a final answer.
    }
    \label{fig:sys_prompt2}
\end{figure}

\paragraph{Synthesis prompts.}
Figures~\ref{fig:sys_prompt1} and~\ref{fig:sys_prompt2} show the prompts used for iterative trajectory synthesis. 
At round $1$, we use \texttt{TOOL\_SELECTION\_PROMPT} and \texttt{TOOL\_CALL\_ONLY\_PROMPT} to separate tool selection from tool execution: the model first selects exactly one tool in the reasoning phase, and then executes the selected tool with concrete parameters without producing a final answer. 
For rounds $>1$, we use \texttt{TOOL\_SELECTION\_OR\_ANSWER\_PROMPT} when no previous answer in the current trajectory has been rejected, allowing the model to either call another tool or provide a final answer based on accumulated observations. 
If an attempted answer is rejected by the verifier, we use \texttt{EXTEND\_TOOL\_SELECTION\_PROMPT} in the following retry to force self-reflection and additional tool use. 
At the maximum round, \texttt{FINAL\_ANSWER\_PROMPT} asks the model to reflect on all observations and produce the final answer.

\begin{figure}[h]
    \centering
    \includegraphics[width=1.0\linewidth]{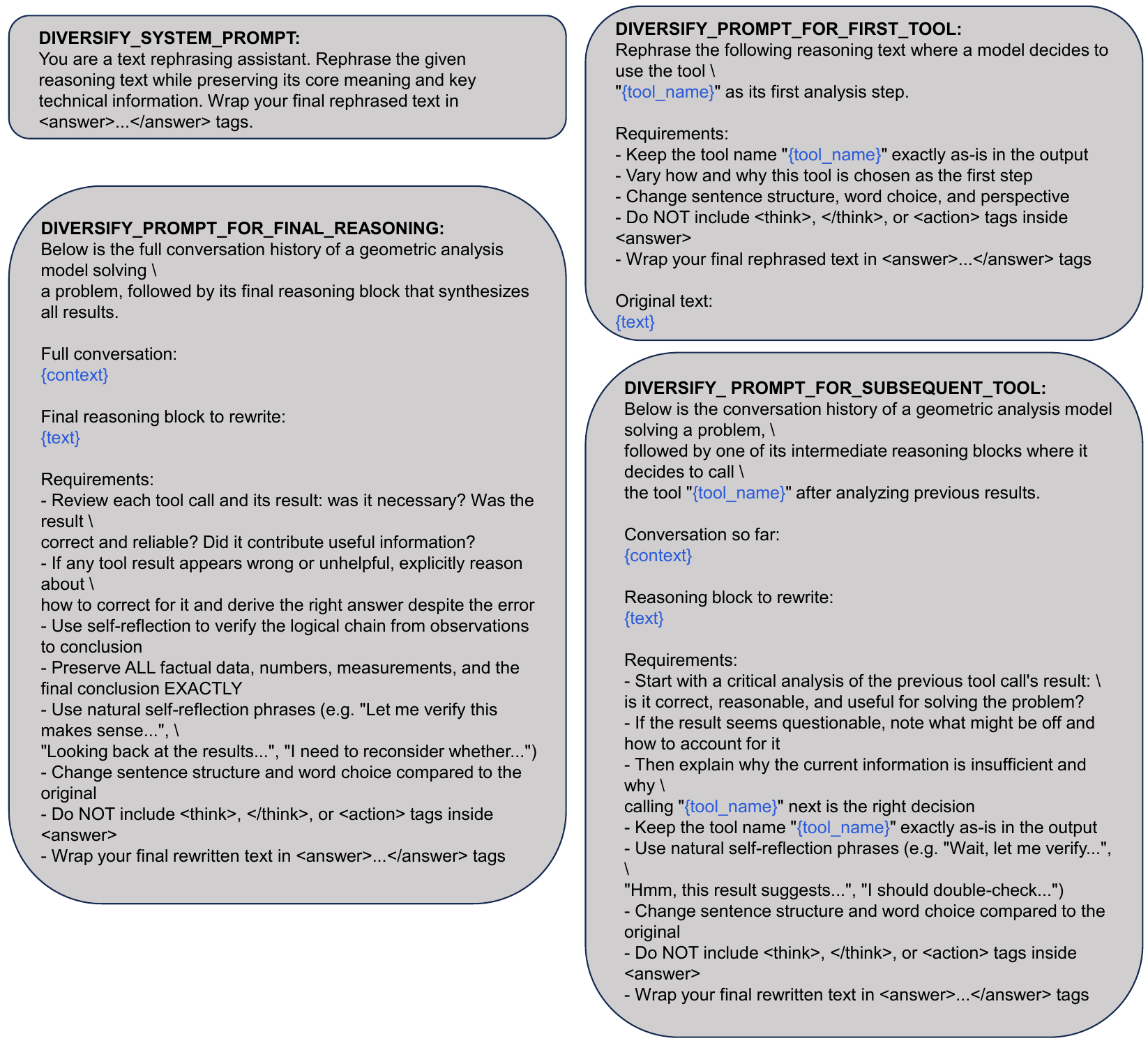}
    \caption{
    \textbf{Diversification rewrites reasoning without changing tool behavior.}
    The prompts rewrite reasoning traces for first-tool selection, subsequent tool selection, and final reasoning while keeping tool calls, observations, and answers fixed.
    }
    \label{fig:diver_prompt}
\end{figure}

\paragraph{Diversification procedure and prompts.}
After collecting valid trajectories, we diversify only the intermediate reasoning text while keeping the tool calls, tool parameters, observations, and final answers unchanged. 
As shown in Figure~\ref{fig:diver_prompt}, \texttt{DIVERSIFY\_SYSTEM\_PROMPT} defines the rephrasing task and asks the model to preserve the core meaning and technical information. 
We then select one of three user prompts according to the role of the \texttt{<think>} block: \texttt{DIVERSIFY\_PROMPT\_FOR\_FIRST\_TOOL} rewrites the first tool-selection reasoning, \texttt{DIVERSIFY\_PROMPT\_FOR\_SUBSEQUENT\_TOOL} rewrites later tool-selection reasoning with conversation context and self-reflection over previous tool results, and \texttt{DIVERSIFY\_PROMPT\_FOR\_FINAL\_REASONING} rewrites the final reasoning and answer block by reviewing all tool calls and preserving all factual data and conclusions. 
The rewritten text is inserted back into the original \texttt{<think>} tags, while outputs containing forbidden tags or missing the required tool name are rejected.

\paragraph{Diversification cost.}\label{app:diversification_cost}
The synthesis set contains 7,661 verified trajectories and 11,767 intermediate \texttt{<think>} blocks.  Rewriting these blocks with Qwen3-VL-235B-A22B-Thinking using TP=8 on one 8$\times$H200 node processes approximately 30M tokens, averages about 150 output tokens per rewrite, and completes in well under eight hours.  Diversification changes only intermediate reasoning text while preserving tool calls, observations, and final answers.

\clearpage

\subsection{OR-GIGPO}
\label{app:alg_details}

\paragraph{Observation-relaxed grouping.}
OR-GIGPO keeps the two-level credit assignment structure described in the main text, but groups actions by their resulting tool observations within the same task rather than by exact pre-action anchor states. 
Semantically equivalent tool calls may produce slightly different outputs, such as reordered OCR text, small detection variations, or different natural-language descriptions from tool agents.

\paragraph{Implementation details.}
For each tool-use turn, we collect the hash and optional text of the observation returned after executing its action, together with the generated token boundary and discounted return. 
We first group turns by the pair of prompt index and observation hash, ensuring that observations are only compared among rollouts from the same multimodal input. 
This exact-hash grouping is applied to all tool observations. 
For image-returning tools, exact matching is usually sufficient because the returned image is deterministic when the tool command and arguments are the same. 
For textual observations returned by tool agents, however, exact matching can be too strict because the same tool behavior may produce minor differences in wording, ordering, or localization descriptions. 
Within each prompt, we further merge exact-hash groups whose representative texts have \texttt{SequenceMatcher} similarity above $\delta=0.9$. 
\texttt{SequenceMatcher} computes a normalized sequence-matching ratio based on the total length of matched subsequences; for two observation texts $a$ and $b$, we use ${\rm sim}(a,b)=2M/(|a|+|b|)$, where $M$ is the total number of matched elements. 
Step-level advantages are computed only for groups with more than one entry, assigned back to the corresponding token spans, and combined with the episode-level advantage using $\omega=1.0$ and $\gamma=0.99$.

\subsection{Rewards}
\label{app:reward_details}

For each rollout $\tau$, we assign the final reward to the last valid response token. 
The trajectory-level reward is defined as:
\[
R(\tau) = R_{\rm rep}(\tau) + R_{\rm format}(\tau) + R_{\rm correct}(\tau).
\]
Here $R_{\rm rep}$ penalizes degenerate repetition, $R_{\rm format}$ encourages valid tool-use trajectories, and $R_{\rm correct}$ measures final-answer correctness.

\paragraph{Repetition penalty.}
We compute the repetition penalty on the model-generated text after removing user turns. 
The penalty is zero for normal responses and becomes negative when repeated patterns are detected:
\begin{itemize}
    \item $-3.0$ if a character is repeated at least 50 times;
    \item $-3.0$ if a word is repeated at least 20 times;
    \item $-2.0$ if a text span of at least 4 words is repeated at least 10 times;
    \item $-1.5 \cdot \frac{2}{T}$ if a sentence is repeated at least 10 times;
    \item $-1.0 \cdot \frac{2}{T}$ if a sentence is repeated at least 7 times;
    \item $-1.5$ if a word is repeated at least 10 times;
    \item $0.0$ otherwise.
\end{itemize}
where $T=\max(\text{number of turns}, 2)$. 
This turn-normalized design avoids over-penalizing longer multi-turn trajectories while discouraging repeated reasoning or tool-use loops.

\paragraph{Format reward.}
The format reward checks whether the trajectory follows the required tool-use protocol:
\begin{itemize}
    \item $+1.0$ if the response contains valid reasoning and a final answer;
    \item $0.0$ if the trajectory contains tool actions but is truncated before the final answer;
    \item $-1.0$ otherwise.
\end{itemize}
A valid trajectory should contain reasoning indicated by \texttt{</think>} and a final answer wrapped by \texttt{<answer>...</answer>}. 
If the response contains \texttt{<action>...</action>} blocks, each action must be preceded by a reasoning segment.

\paragraph{Correctness reward.}
For general spatial QA tasks, $R_{\rm correct}$ follows the evaluation protocol in Appendix~\ref{app:evaluation_details}. 
We first apply rule-based answer matching and then use the LLM-judge fallback described in the evaluation protocol when rule-based matching marks a non-MapTrace prediction as incorrect. 
For non-MapTrace tasks, the final correctness reward is binary:
\[
R_{\rm correct}=1.0 \quad \text{if the prediction is judged correct, and } 0.0 \text{ otherwise.}
\]

\paragraph{MapTrace reward.}
For MapTrace, the $R_{\rm correct}$ uses a continuous NDTW-based score instead of the binary success criterion used in evaluation. 
Given the NDTW distance $d_{\rm NDTW}$, we compute:
\[
R_{\rm correct}^{\rm MapTrace}
=
\begin{cases}
1.0, & d_{\rm NDTW} \leq d_{\rm low}, \\
\dfrac{d_{\rm high}-d_{\rm NDTW}}{d_{\rm high}-d_{\rm low}}, & d_{\rm low} < d_{\rm NDTW} < d_{\rm high}, \\
0.0, & d_{\rm NDTW} \geq d_{\rm high}.
\end{cases}
\]
We set $d_{\rm low}=0.3$ and $d_{\rm high}=0.8$ by default. 
If the predicted path cannot be parsed or the verifier marks it as invalid, we assign $R_{\rm correct}^{\rm MapTrace}=0.0$.

\section{Dataset Details}
\label{app:dataset_details}

\subsection{Training Data}

Table~\ref{tab:training_data_composition} summarizes the data composition used in the SFT and RL stages. 
The main evaluation uses the complete test set of each benchmark; sample-based diagnostics are explicitly identified in their respective sections.

\begin{table}[ht]
    \centering
    \footnotesize
    \setlength{\tabcolsep}{8pt}
    \renewcommand{\arraystretch}{1.1}
    \begin{tabular}{l r r}
        \toprule
        \textbf{Data source} & \textbf{SFT samples} & \textbf{RL samples} \\
        \midrule
        MapTrace & 3,698 & 2,000 \\
        ReasonMap-Plus & 1,979 & 1,580 \\
        MM-MapQA & 1,716 & 2,000 \\
        ReasonMap & 268 & 572 \\
        Mini-o3-Coldstart-Dataset & 3,000 & -- \\
        DeepEyes & -- & 1,500 \\
        Visual Probing & -- & 1,000 \\
        \midrule
        \textbf{Total} & \textbf{10,661} & \textbf{8,652} \\
        \bottomrule
    \end{tabular}
    \caption{
    \textbf{Training combines map reasoning, visual probing, and image-thinking data.}
    We report the data sources and number of samples used in each training stage.
    }
    \label{tab:training_data_composition}
\end{table}

\subsection{Evaluation Domains}

We group the evaluation datasets by task domain to clarify the types of spatial reasoning in our experiments.

\paragraph{Map-based reasoning.}
This group evaluates spatial reasoning over map-like images, where models must connect text labels, landmarks, routes, and global topology. 
\textbf{MapTrace} focuses on route tracing and requires the model to output coordinate paths on maps~\citep{panagopoulou2025maptracescalabledatageneration}. 
\textbf{ReasonMap} evaluates subway-map reasoning over station connectivity, transfer relations, and route topology~\citep{feng2025can}. 
\textbf{ReasonMap-Plus} further expands map reasoning with diverse question types such as counting, true/false verification, and route-related questions~\citep{feng2025rewardmap}. 
\textbf{MapEval} evaluates geospatial map understanding across map-based visual questions~\citep{dihan2025mapevalmapbasedevaluationgeospatial}.

\paragraph{Visual probing and visual search.}
This group evaluates whether models can actively inspect visual evidence and locate task-relevant regions. 
\textbf{Visual Probing} contains visual search and probing tasks that require locating relevant evidence through regional inspection~\citep{lai2025minio3scalingreasoningpatterns}. 
\textbf{V*} focuses on visual search and fine-grained visual reasoning, where the answer often depends on identifying small or localized visual details~\citep{vstar}.

\paragraph{Physical and geometric spatial reasoning.}
This group evaluates spatial transformation, motion, and 3D reasoning beyond map-like domains. 
\textbf{Ball Tracking} tests reasoning over object motion and trajectory changes~\citep{wu2026visual}. 
\textbf{Paper Folding} requires geometric transformation reasoning over folded or unfolded 2D patterns~\citep{wu2026visual}. 
\textbf{Cube Three-View Reasoning} evaluates 3D structure understanding from multiple 2D views~\citep{wu2026visual}. 
\textbf{Real-world Spatial Reasoning} covers spatial relations in natural scenes, requiring models to reason about layouts, object positions, and relative directions~\citep{wu2026visual}.

\paragraph{Basic visual primitives.}
\textbf{BabyVision} evaluates low-level visual reasoning abilities inspired by early human visual development, including visual tracking, spatial perception, fine-grained discrimination, and pattern understanding~\citep{chen2026babyvisionvisualreasoninglanguage}. 
It tests whether models possess robust visual primitives rather than relying only on high-level semantic or language priors.

\subsection{MapTrace Evaluation Criterion Ablation}
\label{sec:maptrace_details}

\begin{table*}[h]
\centering
\footnotesize
\setlength{\tabcolsep}{5.0pt}
\renewcommand\arraystretch{0.98}
\resizebox{0.82\linewidth}{!}{
\begin{tabular}{l | c c c | c}
\toprule
\textbf{Methods ($\downarrow$)}
& \makecell{\textbf{MapTrace}\\\textbf{Thr. 1.0} ($\uparrow$)}
& \makecell{\textbf{MapTrace}\\\textbf{Thr. 0.5} ($\uparrow$)}
& \makecell{\textbf{MapTrace}\\\textbf{Thr. 0.3} ($\uparrow$)}
& \makecell{\textbf{Avg. NDTW}\\($\downarrow$)} \\
\midrule
\multicolumn{5}{l}{\textit{Commercial Models}} \\
\midrule
GPT-5
& \textbf{82.2} & \textbf{61.4} & 41.8 & \textbf{0.720} \\
Gemini-2.5-Flash
& 66.4 & 40.6 & 21.1 & 29.9 \\
Gemini-2.5-Pro
& \underline{80.4} & \underline{60.2} & \underline{42.5} & 3.51 \\

\midrule
\multicolumn{5}{l}{\textit{Open-Source VLMs and Tool/Reason-Augmented Baselines}} \\
\midrule
InternVL3.5-8B
& 70.8 & 48.7 & 31.0 & 5.92 \\
InternVL3.5-14B
& 76.2 & 55.8 & 37.1 & 3.27 \\
Qwen3-30B-A3B
& 75.0 & 55.0 & 39.1 & 0.796 \\
Qwen3-32B
& 75.5 & 53.7 & 35.5 & 0.811 \\
Qwen3-235B
& 72.9 & 53.5 & 37.5 & 0.827 \\
VTool-R1-3B
& 40.6 & 22.8 & 13.2 & 2.14 \\
VTool-R1-7B
& 45.2 & 29.8 & 19.4 & 28.3 \\
R1-OneVision-7B
& 41.9 & 22.9 & 13.3 & 1012 \\
Mini-o3
& 30.4 & 12.8 & 6.2 & 3.33 \\

\midrule
\multicolumn{5}{l}{\textit{Our Method}} \\
\midrule
Qwen3-2B
& 66.2 & 45.8 & 30.7 & 3.56 \\
\rowcolor{gray!15}
\method-2B
& 72.2\gain{6.0}
& 56.2\gain{10.4}
& 38.4\gain{7.7}
& 0.812{\tiny\textcolor{green!50!black}{ $-2.74$}} \\
Qwen3-4B
& 71.6 & 49.9 & 34.0 & 0.902 \\
\rowcolor{gray!15}
\method-4B
& 74.5\gain{2.9}
& 57.9\gain{8.0}
& 40.7\gain{6.7}
& 0.818{\tiny\textcolor{green!50!black}{ $-0.084$}} \\
Qwen3-8B
& 74.1 & 57.5 & 38.4 & 0.822 \\
\rowcolor{gray!15}
\method-8B (ours)
& 77.7\gain{3.6}
& 59.2\gain{1.7}
& \textbf{44.1}\gain{5.7}
& \underline{0.778}{\tiny\textcolor{green!50!black}{ $-0.044$}} \\
\bottomrule
\end{tabular}%
}
\caption{
\textbf{MapTrace evaluation under different NDTW thresholds.}
We report threshold-based accuracy under three NDTW thresholds and the average NDTW distance.
Higher is better for threshold-based accuracy, while lower is better for Avg. NDTW. Accuracy gains are percentage points; NDTW changes are distance differences.
\method-8B improves over the Qwen3-8B backbone across all threshold settings and achieves the best accuracy among all models under the strictest threshold. Bold and underlined values indicate the best and second-best results across all listed models, respectively.
}
\label{tab:maptrace_threshold}
\end{table*}

We use the NDTW path distance from MapTrace~\citep{panagopoulou2025maptracescalabledatageneration}, but summarize it as threshold-based accuracy rather than Avg. NDTW: for a threshold $\tau$, we report the fraction of test examples with distance below $\tau$. The main results use $\tau=1.0$, while Table~\ref{tab:maptrace_threshold} also reports $\tau=0.5$ and $0.3$. 
As shown in Table~\ref{tab:maptrace_threshold}, Avg. NDTW is sensitive to extreme values and may not reliably reflect the capability of different models. 
We therefore also report model performance under stricter thresholds. 
\method-8B achieves the highest threshold-based accuracy among non-commercial models at all three thresholds, with a larger gain over its backbone under the strictest criterion. 
At the threshold of $0.3$, \method-8B achieves the best result among all models.

\section{Additional Experiments}
\label{app:additional_experiments}

This appendix reports additional experiments on the training recipe of \method.  Appendix~\ref{app:teacher_quality} examines how the quality of the trajectory teacher affects SFT, and Appendix~\ref{app:group_size} studies the rollout group size of OR-GIGPO.

\subsection{Teacher Quality}
\label{app:teacher_quality}
We replaced the GPT-5 trajectory teacher with the open-weight Qwen3.5-397B-A17B teacher while keeping the diversification model, SFT budget, and evaluation protocol fixed.  Each SFT-only checkpoint was evaluated once on the same stratified 10\% sample from all 13 benchmarks.  Table~\ref{tab:additional_teacher} shows that the open-weight teacher already improves over the no-SFT checkpoint, while GPT-5 provides a further gain; this is a diagnostic of teacher quality rather than a replacement for the full-test-set main comparison.

\begin{table}[!htbp]
    \centering
    \small
    \setlength{\tabcolsep}{5pt}
    \begin{tabular}{lc}
        \toprule
        \textbf{SFT setting} & \textbf{Average accuracy} \\
        \midrule
        Qwen3-VL-8B, no SFT & 34.1 \\
        Qwen3.5-397B-A17B teacher & 37.6 \\
        GPT-5 teacher & \textbf{39.8} \\
        \bottomrule
    \end{tabular}
    \caption{Teacher-quality diagnostic on the same stratified 10\% sample from all 13 benchmarks.  The open-weight teacher improves over the no-SFT checkpoint, while GPT-5 provides a further gain in this diagnostic.}
    \label{tab:additional_teacher}
\end{table}

\subsection{Rollout Group Size}
\label{app:group_size}
For the rollout-group-size diagnostic, we evaluated Qwen3-VL-2B-Thinking with a fixed 10\% sample from each of the six in-distribution benchmarks, shared by all settings in this comparison.  As shown in Table~\ref{tab:additional_group_size}, increasing the group size from $N=4$ to $N=8$ yields a further improvement within the tested range.

\begin{table}[!htbp]
    \centering
    \small
    \setlength{\tabcolsep}{5pt}
    \begin{tabular}{lcc}
        \toprule
        \textbf{Setting} & \textbf{Average accuracy} & \textbf{Gain vs. backbone (pp)} \\
        \midrule
        Qwen3-VL-2B backbone & 22.6 & -- \\
        OR-GIGPO ($N=4$) & 36.9 & +14.3 \\
        OR-GIGPO ($N=8$) & \textbf{38.2} & \textbf{+15.6} \\
        \bottomrule
    \end{tabular}
    \caption{Completed rollout-group-size diagnostic on a fixed 10\% sample from each of the six in-distribution benchmarks.  Within this tested range, $N=8$ adds 1.3 percentage points over $N=4$.}
    \label{tab:additional_group_size}
\end{table}

\section{Fine-grained Tool-use Analysis}
\label{app:tool_analysis}

This appendix analyzes how \method\ uses its tools.  Appendix~\ref{app:tool_distribution} reports the tool-call distribution under different tool conditions, Appendix~\ref{app:interface_transfer} evaluates whether prior agents benefit from the \method\ sandbox without retraining, Appendix~\ref{app:map_tools} examines the role of map-oriented tools and compares map and non-map benchmarks, and Appendix~\ref{app:cube} analyzes tool effectiveness on Cube Three-View.

\subsection{Tool-use Distribution}
\label{app:tool_distribution}

\begin{figure}[h]
\centering
\includegraphics[width=\linewidth]{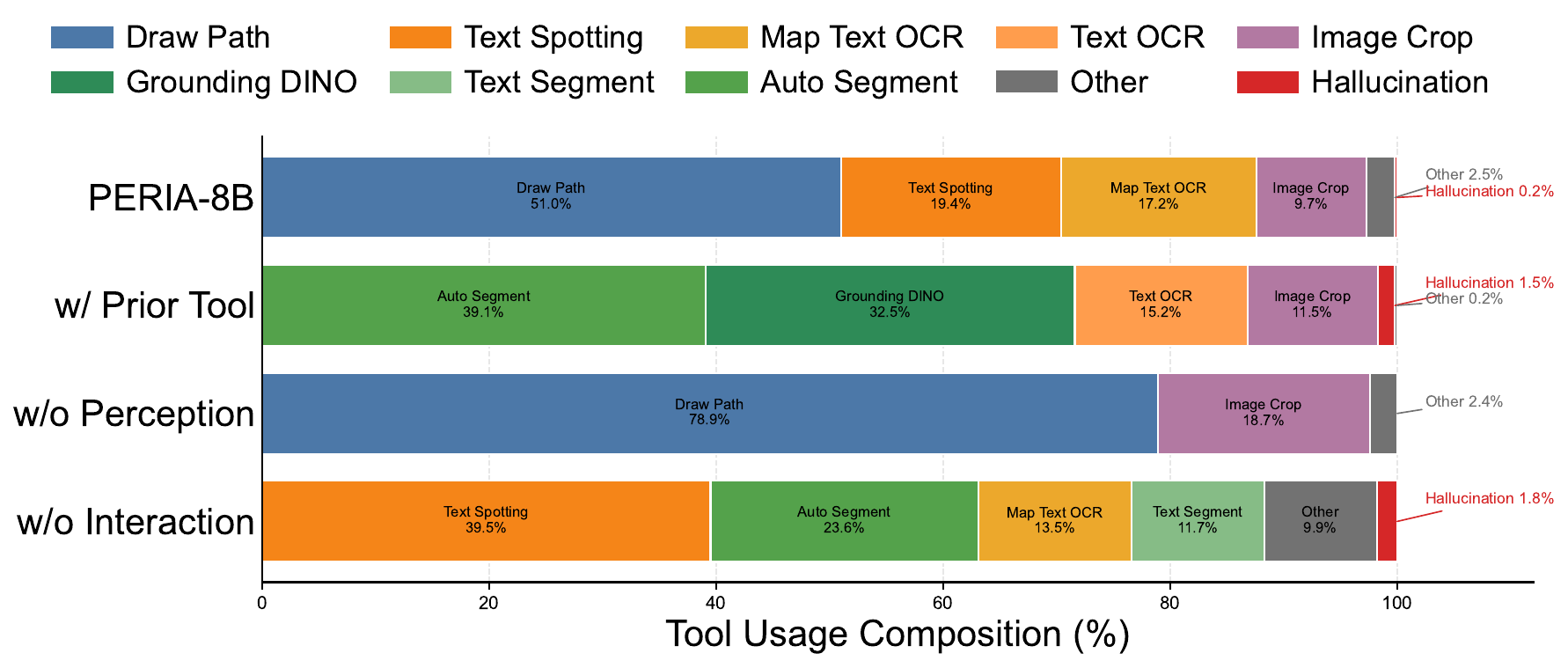}
\caption{
\textbf{Tool-use distribution under different inference tool conditions.}
\method-8B adapts to different available tool sets by using diverse tools.
Compared with the previous-tool setting, the full sandbox shows a substantially different usage pattern, suggesting that newly designed tools are actively used for reasoning.
}
\label{fig:tool_use_distribution}
\end{figure}

We further analyze how \method-8B uses different tools under the inference-time tool conditions in Table~\ref{tab:ablation}. Figure~\ref{fig:tool_use_distribution} reports the normalized tool-call composition for the full sandbox, the previous-tool setting, the setting without perception tools, and the setting without interaction tools. With the full 18-tool sandbox, \method-8B primarily uses \texttt{draw\_path} ($51.0\%$), \texttt{text\_spotting} ($19.4\%$), \texttt{map\_text\_ocr} ($17.2\%$), and \texttt{image\_crop} ($9.7\%$), while hallucinated tool calls remain rare ($0.2\%$). This distribution shows that the trained agent actively combines interaction tools for path-level reasoning with perception tools for extracting textual and spatial evidence.

Under the previous-tool setting, where only generic tools from prior work are available, the usage pattern changes substantially: the model relies mainly on \texttt{auto\_segment} ($39.1\%$), \texttt{grounding\_dino} ($32.5\%$), \texttt{text\_ocr} ($15.2\%$), and \texttt{image\_crop} ($11.5\%$). This shift suggests that the full \method\ sandbox does not merely increase the number of available tools, but introduces spatially targeted tools that are preferred by the trained agent and lead to stronger performance. In particular, the frequent use of \texttt{draw\_path}, \texttt{text\_spotting}, and \texttt{map\_text\_ocr} in the full setting aligns with the needs of route tracing, map reasoning, and fine-grained spatial verification.

The constrained settings further reveal the complementary roles of the two tool families. When perception tools are removed, the model reallocates calls to interaction tools, especially \texttt{draw\_path} ($78.9\%$) and \texttt{image\_crop} ($18.7\%$). When interaction tools are removed, it instead relies on perception tools such as \texttt{text\_spotting} ($39.5\%$), \texttt{auto\_segment} ($23.6\%$), \texttt{map\_text\_ocr} ($13.5\%$), and \texttt{text\_segment} ($11.7\%$). These reallocations show that \method-8B can adapt to different tool availability, but the corresponding performance drops in Table~\ref{tab:ablation} indicate that either tool family alone is insufficient. We also observe slightly higher hallucinated tool-call rates under constrained settings, especially when interaction tools are removed, suggesting that restricting the sandbox makes tool selection harder. Overall, the analysis supports the effectiveness of our tool design and the necessity of coordinating perception and interaction tools for robust spatial reasoning.

\subsection{Zero-shot Interface Transfer}
\label{app:interface_transfer}
To test whether access to the PERIA sandbox alone explains the performance gap, we evaluated released checkpoints with the PERIA inference prompt and sandbox without retraining.  All models were evaluated on the same stratified 10\% sample from all 13 benchmarks.  The native column uses each checkpoint's original interface; the PERIA-sandbox column uses the transferred interface.  Table~\ref{tab:additional_transfer} shows that sandbox access alone does not reproduce the gains of \method, indicating that tool competence must be learned.

\begin{table}[!htbp]
    \centering
    \small
    \setlength{\tabcolsep}{4pt}
    \begin{tabular}{lcc}
        \toprule
        \textbf{Model} & \textbf{Native} & \textbf{PERIA sandbox} \\
        \midrule
        VTool-R1-7B & 34.5 & 35.0 \\
        R1-OneVision-7B & 27.4 & 29.4 \\
        Mini-o3 & 30.7 & 13.1 \\
        \method-8B & -- & \textbf{51.7} \\
        \bottomrule
    \end{tabular}
    \caption{Zero-shot interface transfer on a common stratified 10\% sample from all 13 benchmarks.  Sandbox access alone gives small changes for the prior agents and substantially reduces Mini-o3 performance; the transferred values do not constitute retrained baselines.}
    \label{tab:additional_transfer}
\end{table}

\subsection{Map-oriented Tools and Map/Non-map Balance}
\label{app:map_tools}

We evaluated the trained checkpoint with both \texttt{map\_text\_ocr} and \texttt{draw\_path} removed at inference time on the complete test sets of ReasonMap-Plus, ReasonMap, and MapTrace.  This is a full-test-set diagnostic for these three map benchmarks only.  As shown in Table~\ref{tab:additional_map_tools}, the restricted model remains above the backbone on all three tasks.

\begin{table}[!htbp]
    \centering
    \small
    \setlength{\tabcolsep}{3pt}
    \begin{tabular}{lcccc}
        \toprule
        \textbf{Setting} & \textbf{ReasonMap-Plus} & \textbf{ReasonMap} & \textbf{MapTrace} & \textbf{Avg.} \\
        \midrule
        Qwen3-VL-8B & 65.7 & 12.3 & 74.1 & 50.7 \\
        \method-8B, all tools & 72.6 & \textbf{27.1} & \textbf{77.7} & \textbf{59.1} \\
        \method-8B, w/o map tools & \textbf{73.4} & 18.3 & 74.7 & 55.5 \\
        \bottomrule
    \end{tabular}
    \caption{Map-oriented-tool removal on complete test sets of three map benchmarks.  The restricted model remains above the backbone on all three tasks, while losing 3.7 percentage points before rounding relative to the full-tool setting.}
    \label{tab:additional_map_tools}
\end{table}

Across the four map-related benchmarks, the reported average increases from 51.5\% for the backbone to 58.2\% for \method, a gain of 6.7 percentage points.  Across the nine non-map benchmarks, the corresponding average increases from 31.2\% to 38.3\%, a gain of 7.1 percentage points.  These aggregate comparisons use the main Table~\ref{tab:main_results} values; they summarize benchmark groups and do not introduce a new sampling protocol.

\subsection{Tool Effectiveness on Cube Three-View}
\label{app:cube}
On a 10\% sample of the held-out Cube Three-View test set, \method-8B obtains 35.4\% with tools enabled and 22.9\% with direct chain-of-thought without tools.  At least one tool is invoked on 75\% of examples, most often \texttt{text\_spotting}.  The tool-enabled/direct-CoT comparison is sample-based.  Separately, excluding Cube Three-View from the main Table~\ref{tab:main_results} leaves a 2.6-percentage-point average improvement over the backbone across the remaining six OOD benchmarks.

\section{Qualitative Examples}
\label{app:case_studies}

\begin{figure}[h]
    \centering
    \includegraphics[width=0.95\linewidth]{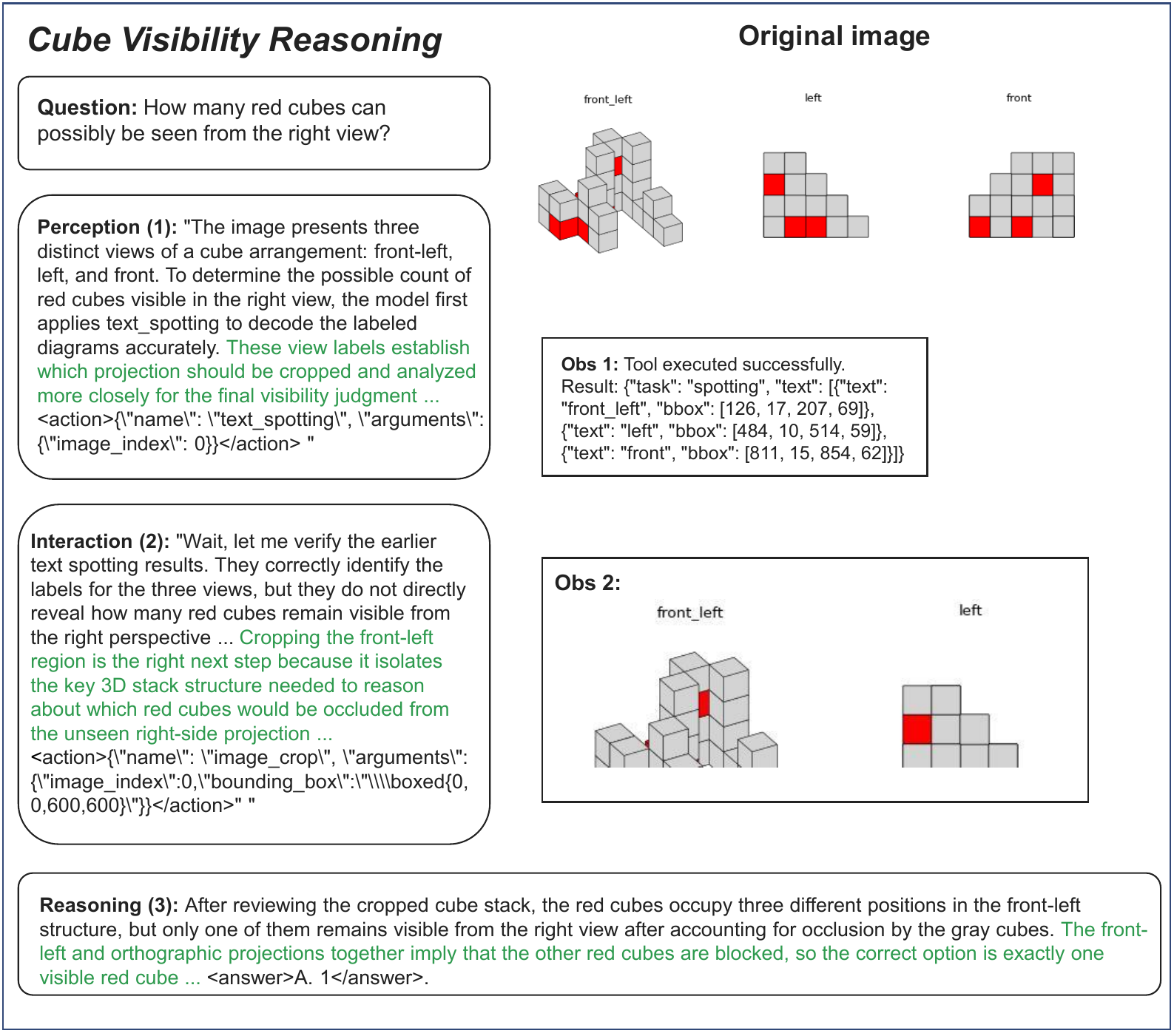}
    \caption{
    Qualitative example of cube visibility reasoning.
    Text spotting identifies the view labels, while the final answer requires cropping the cube arrangement and reasoning about 3D occlusion.
    }
    \label{fig:qualitative_cube_visibility}
\end{figure}

\paragraph{Example: From view identification to spatial reasoning.}
Figure~\ref{fig:qualitative_cube_visibility} shows an example that requires reasoning over multiple cube views.
The question asks how many red cubes can possibly be seen from the right view.
The model first uses text spotting to identify the three view labels: ``front\_left'', ``left'', and ``front''.
These labels help ground the diagrams, but they do not directly answer the visibility question.

To obtain the needed visual evidence, the model crops the front-left region and inspects the 3D cube structure.
Based on the relative positions of the red and gray cubes, it infers that only one red cube remains visible from the right view after occlusion.
The final answer is \texttt{A. 1}.
This example shows that view labels are useful for grounding, while the final answer still requires localized spatial reasoning.

\begin{figure}[t]
    \centering
    \includegraphics[width=0.95\linewidth]{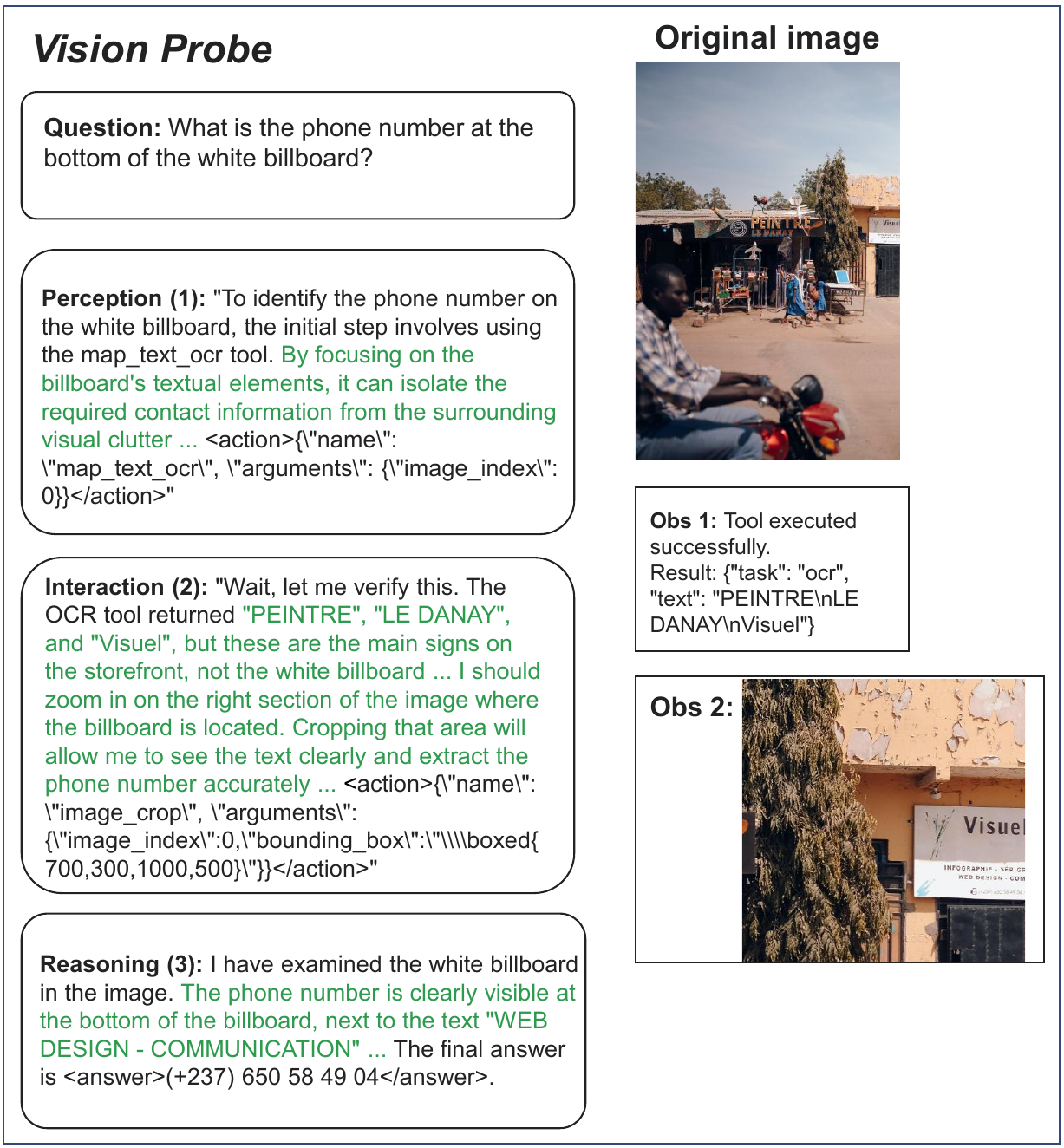}
    \caption{
    Qualitative example of iterative vision probing.
    The OCR output contains visible storefront text but omits the phone number required by the question.
    Cropping the relevant billboard region provides the missing evidence.
    }
    \label{fig:qualitative_ocr_recovery}
\end{figure}

\paragraph{Example: Recovering from incomplete OCR evidence.}
Figure~\ref{fig:qualitative_ocr_recovery} shows an example where the initial OCR output is correct but incomplete for the question.
The question asks for the phone number at the bottom of a white billboard.
The model first applies OCR to the full image, which returns visible storefront text such as ``PEINTRE'', ``LE DANAY'', and ``Visuel''.
However, the returned text does not include the queried phone number.

The model therefore checks the sufficiency of the OCR evidence and crops the right side of the image where the billboard appears.
After this localized inspection, the phone number becomes visible, leading to the answer \texttt{(+237) 650 58 49 04}.
This example shows that a tool output can be useful but non-exhaustive, and that additional localization can recover missing task-relevant evidence.

Together, these examples show that intermediate tool outputs should be checked for sufficiency, and that perception and interaction tools play complementary roles. 
Perception tools provide global cues such as text labels and OCR evidence, while interaction tools enable localized inspection through cropping or other visual operations. 
When the returned evidence is incomplete or only partially relevant, combining these two tool families helps the agent obtain the information needed for the final answer.

\clearpage
\section*{NeurIPS Paper Checklist}

\begin{enumerate}

\item {\bf Claims}
    \item[] Question: Do the main claims made in the abstract and introduction accurately reflect the paper's contributions and scope?
    \item[] Answer: \answerYes{} % Replace by \answerYes{}, \answerNo{}, or \answerNA{}.
    \item[] Justification: The abstract and introduction focus on the \method\ framework, perception--interaction tool sandbox, trajectory synthesis, OR-GIGPO training, and empirical gains across spatial reasoning benchmarks, all of which are directly supported by the methodology, ablations, and experimental results.

\item {\bf Limitations}
    \item[] Question: Does the paper discuss the limitations of the work performed by the authors?
    \item[] Answer: \answerYes{} % Replace by \answerYes{}, \answerNo{}, or \answerNA{}.
    \item[] Justification: The paper discusses limitations and future work in Appendix~\ref{app:limit_future}, including training scale, model scale, inference-turn limits, and extensions to broader spatial reasoning domains and modalities.

\item {\bf Theory assumptions and proofs}
    \item[] Question: For each theoretical result, does the paper provide the full set of assumptions and a complete (and correct) proof?
    \item[] Answer: \answerNA{} % Replace by \answerYes{}, \answerNo{}, or \answerNA{}.
    \item[] Justification: The paper does not present theoretical results requiring formal assumptions or proofs; it focuses on method design, training algorithms, and empirical evaluation.

    \item {\bf Experimental result reproducibility}
    \item[] Question: Does the paper fully disclose all the information needed to reproduce the main experimental results of the paper to the extent that it affects the main claims and/or conclusions of the paper (regardless of whether the code and data are provided or not)?
    \item[] Answer: \answerYes{} % Replace by \answerYes{}, \answerNo{}, or \answerNA{}.
    \item[] Justification: The paper provides the model backbones, evaluation datasets and metrics, baseline settings, training recipe, SFT and RL hyperparameters, tool sandbox details, reward design, OR-GIGPO implementation details, prompt templates, and inference settings in the main text and appendices, providing a reasonable path to reproduce or verify the main experimental results.

\item {\bf Open access to data and code}
    \item[] Question: Does the paper provide open access to the data and code, with sufficient instructions to faithfully reproduce the main experimental results, as described in supplemental material?
    \item[] Answer: \answerYes{} % Replace by \answerYes{}, \answerNo{}, or \answerNA{}.
    \item[] Justification: The public code and data links are provided on the first page. The code repository includes the tool sandbox, training and evaluation scripts, prompt templates, and instructions for data preparation, training, and inference.

\item {\bf Experimental setting/details}
    \item[] Question: Does the paper specify all the training and test details (e.g., data splits, hyperparameters, how they were chosen, type of optimizer) necessary to understand the results?
    \item[] Answer: \answerYes{} % Replace by \answerYes{}, \answerNo{}, or \answerNA{}.
    \item[] Justification: The paper describes the evaluation datasets, in-distribution and out-of-distribution splits, metrics, baselines, model backbones, SFT and RL training settings, and inference setup in the main text, with additional hyperparameters, prompts, tool settings, and implementation details provided in the appendix.

\item {\bf Experiment statistical significance}
    \item[] Question: Does the paper report error bars suitably and correctly defined or other appropriate information about the statistical significance of the experiments?
    \item[] Answer: \answerNo{} % Replace by \answerYes{}, \answerNo{}, or \answerNA{}.
    \item[] Justification: The paper reports deterministic benchmark accuracy results and extensive ablations, but does not include error bars, confidence intervals, or statistical significance tests. We acknowledge this as a limitation and plan to include repeated-run statistics in future versions.

\item {\bf Experiments compute resources}
    \item[] Question: For each experiment, does the paper provide sufficient information on the computer resources (type of compute workers, memory, time of execution) needed to reproduce the experiments?
    \item[] Answer: \answerNo{} % Replace by \answerYes{}, \answerNo{}, or \answerNA{}.
    \item[] Justification: The appendix reports hardware and runtime information for reasoning diversification, but per-experiment training and inference runtimes, total compute, and compute used for preliminary experiments are not fully reported.
    
\item {\bf Code of ethics}
    \item[] Question: Does the research conducted in the paper conform, in every respect, with the NeurIPS Code of Ethics \url{https://neurips.cc/public/EthicsGuidelines}?
    \item[] Answer: \answerYes{} % Replace by \answerYes{}, \answerNo{}, or \answerNA{}.
    \item[] Justification: The research is conducted on visual reasoning benchmarks and tool-use training/evaluation without involving human subjects, private personal data, or deceptive data collection, and we have reviewed the NeurIPS Code of Ethics to ensure compliance.

\item {\bf Broader impacts}
    \item[] Question: Does the paper discuss both potential positive societal impacts and negative societal impacts of the work performed?
    \item[] Answer: \answerYes{} % Replace by \answerYes{}, \answerNo{}, or \answerNA{}.
    \item[] Justification: The paper discusses broader impacts in the Conclusion, including the potential benefits of spatial reasoning agents for visual assistance, mapping, and planning, as well as possible risks from incorrect tool-grounded reasoning, over-reliance on agent outputs, and misuse in surveillance or navigation-sensitive settings.
    
\item {\bf Safeguards}
    \item[] Question: Does the paper describe safeguards that have been put in place for responsible release of data or models that have a high risk for misuse (e.g., pre-trained language models, image generators, or scraped datasets)?
    \item[] Answer: \answerNA{} % Replace by \answerYes{}, \answerNo{}, or \answerNA{}.
    \item[] Justification: The paper does not release high-risk generative models, scraped sensitive datasets, or data intended for surveillance or identity-related applications. The planned release focuses on tool-use training code, tool sandbox components, prompts, and spatial reasoning trajectories for benchmark-style tasks.

\item {\bf Licenses for existing assets}
    \item[] Question: Are the creators or original owners of assets (e.g., code, data, models), used in the paper, properly credited and are the license and terms of use explicitly mentioned and properly respected?
    \item[] Answer: \answerYes{} % Replace by \answerYes{}, \answerNo{}, or \answerNA{}.
    \item[] Justification: All external models, datasets, and tools used in the paper are properly cited with references to their original publications, and their licenses or terms of use are acknowledged where applicable. This includes VLM backbones, prior tool sets, and benchmark datasets.

\item {\bf New assets}
    \item[] Question: Are new assets introduced in the paper well documented and is the documentation provided alongside the assets?
    \item[] Answer: \answerYes{} % Replace by \answerYes{}, \answerNo{}, or \answerNA{}.
    \item[] Justification: The paper introduces new assets including the \method\ tool sandbox, synthesized tool-use trajectories, prompts, and training/evaluation code. These assets are documented in the main text and appendices, including tool definitions, trajectory synthesis procedures, reward details, prompts, implementation settings, and limitations, and will be released with accompanying documentation.

\item {\bf Crowdsourcing and research with human subjects}
    \item[] Question: For crowdsourcing experiments and research with human subjects, does the paper include the full text of instructions given to participants and screenshots, if applicable, as well as details about compensation (if any)? 
    \item[] Answer: \answerNA{} % Replace by \answerYes{}, \answerNo{}, or \answerNA{}.
    \item[] Justification: The paper does not involve crowdsourcing experiments or research with human subjects.

\item {\bf Institutional review board (IRB) approvals or equivalent for research with human subjects}
    \item[] Question: Does the paper describe potential risks incurred by study participants, whether such risks were disclosed to the subjects, and whether Institutional Review Board (IRB) approvals (or an equivalent approval/review based on the requirements of your country or institution) were obtained?
    \item[] Answer: \answerNA{} % Replace by \answerYes{}, \answerNo{}, or \answerNA{}.
    \item[] Justification: The paper does not involve crowdsourcing experiments or research with human subjects, so IRB approval or equivalent review is not applicable.

\item {\bf Declaration of LLM usage}
    \item[] Question: Does the paper describe the usage of LLMs if it is an important, original, or non-standard component of the core methods in this research? Note that if the LLM is used only for writing, editing, or formatting purposes and does \emph{not} impact the core methodology, scientific rigor, or originality of the research, declaration is not required.
    \item[] Answer: \answerYes{} % Replace by \answerYes{}, \answerNo{}, or \answerNA{}.
    \item[] Justification: The paper explicitly describes the use of LLMs/VLMs in the core methodology, including GPT-5 for tool-use trajectory synthesis, Qwen3-VL-235B-A22B-Thinking for reasoning-style diversification, and an LLM-judge fallback for correctness verification in reward computation.

\end{enumerate}

\end{document}